\documentclass[lettersize,journal,twoside, 10pt]{IEEEtran}  
                               
\title{\LARGE \bf
An Object Deformation-Agnostic Framework for Human-Robot Collaborative Transportation}

\author{Doganay Sirintuna~\IEEEmembership{Member, IEEE}, Alberto Giammarino~\IEEEmembership{Member, IEEE}, and Arash Ajoudani~\IEEEmembership{Member, IEEE}
\thanks{The authors are with the HRI$^{2}$ Lab, Istituto Italiano di Tecnologia, Genoa, Italy.
        {\tt\small \{doganay.sirintuna\}@iit.it}}
\thanks{This work was supported in part by the ERC-StG Ergo-Lean (Grant Agreement No.850932), in part by the European Union’s Horizon 2020 research and innovation programme under Grant Agreement No. 871237 (SOPHIA) and No. 101016007 (CONCERT).}
\thanks{The authors thank to Dr. Marta Lorenzini, Mattia Leonori and Dr. Juan Manuel Gandarias Palacios for the fruitful discussions.}}

\usepackage{cite}
\usepackage{amsmath}
\usepackage{amssymb}
\usepackage[pdftex]{graphicx}
\usepackage[dvipsnames]{xcolor}
\usepackage[hyphens]{url}
\usepackage{caption}
\usepackage{algorithm}
\usepackage{algpseudocode}
\usepackage[english]{datetime2}

\DeclareCaptionFont{mysize}{\fontsize{8}{9.6}\selectfont}
\begin{document}

\maketitle

\begin{abstract}

In this study, an adaptive object deformability-agnostic human-robot collaborative transportation framework is presented. The proposed framework enables to combine the haptic information transferred through the object with the human kinematic information obtained from a motion capture system to generate reactive whole-body motions on a mobile collaborative robot. Furthermore, it allows rotating the objects in an intuitive and accurate way during co-transportation based on an algorithm that detects the human rotation intention using the torso and hand movements. First, we validate the framework with the two extremities of the object deformability range (i.e, purely rigid aluminum rod and highly deformable rope) by utilizing a mobile manipulator which consists of an Omni-directional mobile base and a collaborative robotic arm. Next, its performance is compared with an admittance controller during a co-carry task of a partially deformable object in a 12-subjects user study. Quantitative and qualitative results of this experiment show that the proposed framework can effectively handle the transportation of objects regardless of their deformability and provides intuitive assistance to human partners. Finally, we have demonstrated the potential of our framework in a different scenario, where the human and the robot co-transport a manikin using a deformable sheet. 

\end{abstract}

\def\abstractname{Note to Practitioners}
\begin{abstract}
    Transportation of objects which requires the cooperation of multiple partners is a common task in industrial settings such as factories and warehouses. The existing human-robot collaboration solutions for this task have focused only on purely rigid objects, although deformable objects need to be carried frequently in real-world applications. In this paper, we introduce a human-robot collaborative transportation framework that can handle objects with different deformability ranging from purely rigid to highly deformable.  In particular, the proposed framework generates whole-body movements on a mobile collaborative robot by using the combination of the haptic information transmitted through the object and the human motion information obtained from a motion capture system. Moreover, the framework includes an intuitive way to rotate the object during the execution based on human hand and torso motion. The results of the experiments where objects with various deformability characteristics were transported in collaboration with a mobile manipulator demonstrated the high potential of the proposed approach in a laboratory setting. In the future, we plan to employ a less expensive vision-based human motion tracking system instead of the IMU-based system used in this study. With this change, we will be able to eliminate the need for wearable sensors from the framework presented, which would enhance its usability in real-world scenarios.
\end{abstract}

\begin{IEEEkeywords}
Physical Human-Robot Interaction, Collaborative Object Transportation, Deformable Materials, Mobile Manipulation, Whole-Body Control.
\end{IEEEkeywords}

\begin{figure}[ht!]
    \centering
    \resizebox{1\columnwidth}{!}{\rotatebox{0}{\includegraphics[trim=0cm 0cm 0cm 0cm, clip=true]{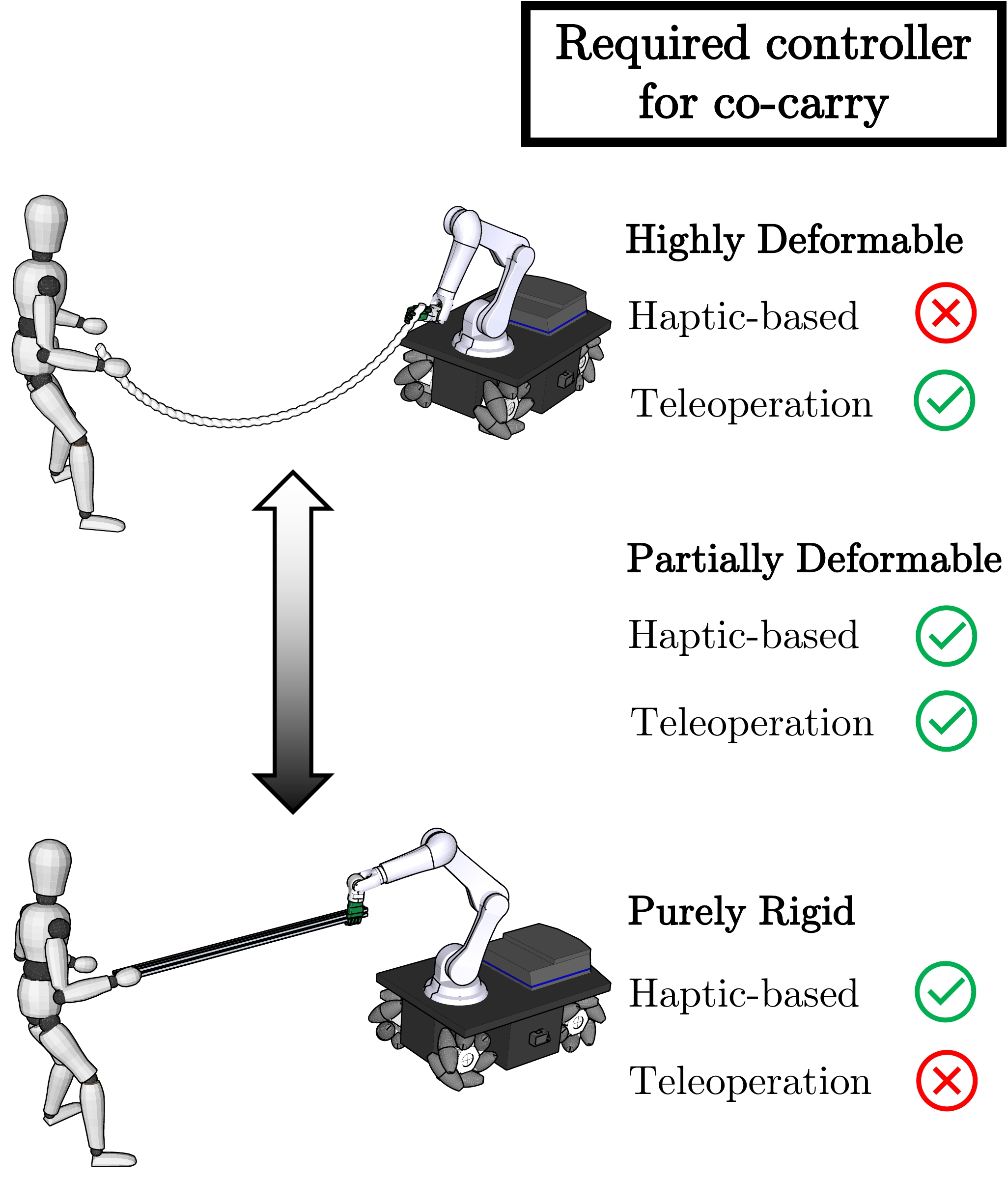}}}
    \caption{On the left, human-robot co-transportation of objects having different deformation characteristics, ranging from highly deformable (e.g., a rope) to rigid (e.g., an aluminum profile). On the right, the controller type that can deal with the corresponding object deformability. Haptic-based controllers solely use haptic feedback to compute the robot desired motion, while teleoperation frameworks utilize human body movements. We propose an adaptive control framework that merges haptic and human motion information to enable collaborative carrying objects irrespective of their deformation characteristics.}
    \label{fig:digest_figure}
    \vspace{-0.0cm}
\end{figure}

\section{INTRODUCTION}
\label{sec:introduction}

Industrial automation is a crucial enabler for improving productivity and quality, while reducing errors and waste, and for preparing a response to the rapidly aging workforce~\cite{krueger2017have}. In this direction, robotic technologies have been increasingly adopted to replace the human workforce in several repetitive and simple operations. Nevertheless, robots of today still lack operational flexibility and intelligence, typical of humans, which limits their applicability in industrial environments where high flexibility is also required (e.g., logistics). Consequently, workers are still involved in physically demanding tasks, that might jeopardize their health and productivity~\cite{dework}. Building robotic technologies that can effectively collaborate with humans represents one viable way to tackle these issues~\cite{kim2019adaptable}.

One frequently encountered task in industrial settings such as factories, warehouses, and construction sites is the transportation of objects. In some cases, this task can be physically demanding while requiring a certain level of adaptability to the surroundings. Moreover, it often requires the cooperation of multiple partners. Instead of an automated solution to this problem, collaborative robots can be used to work together with humans to exploit their cognitive skills in unstructured environments~\cite{Arash,gandarias2022enhancing}.

Three main challenges can be identified in the literature on human-robot co-transportation of objects. First, the robot should effectively share the load, without hindering human intention~\cite{mortl2012role}. Since humans can easily lead the motion during collaboration, the robot should be capable of effectively following them. The second one is the well-known translation/rotation ambiguity~\cite{dumora2012experimental}. Especially, when long objects are carried, humans can hardly control both rotations and translations independently while grasping only one side of the object. This is due to the coupling between different degrees of freedom (DoFs) in the object dynamics~\cite{Takubo2002}. Thirdly, co-transportation of deformable objects represents a major challenge in this research field. Due to the incomplete wrenches transmitted through the object, most haptic-based conventional techniques fall short in this condition.

In this article, we present a framework for human-robot co-transportation aiming at tackling the three challenges previously mentioned. In order to validate our framework, we use a mobile base robotic platform, which allows movements over a large workspace. It consists of a $n_b$ DoFs Omni-directional mobile base, a $n_a$ DoFs robotic arm with a Force/Torque (F/T) sensor at its end-effector to measure the wrenches applied throughout the interaction. In addition, a motion capture (MoCap) system is employed to track movements of the human operator during the co-carrying task. The sensory systems' choice is mainly based on the acquisition of adequate information to enable co-carrying of objects with different deformability (from highly deformable to purely rigid) or those with different perceived deformability in different directions of co-carrying (e.g., a rope that can be relatively rigid when pulled in the constrained direction, generating pure wrenches, but very loose in the opposite direction, generating pure twists). In particular, the contributions of this paper can be considered as three-fold:

\begin{itemize}
    \item The design of an adaptive framework that merges haptic and human movement information to generate motion references for a mobile manipulator, enabling it to co-transport objects independently of their deformability during collaboration with a human (Fig.~\ref{fig:digest_figure}).
    \item The development of an algorithm that enables human partners to convey their intention of rotating an object to a desired pose in an intuitive and accurate way.
    \item The experimental validation of the developed system that includes a comparison with control frameworks which are limited to the use of either haptic or human motion information in the extremities of the object deformability range, a multi-subject cross-gender user study with the obtained quantitative and qualitative results, and a practical showcase where the applicability of our framework has been tested in a challenging scenario.
\end{itemize}

The original idea of this study was presented in our previous conference paper~\cite{doganay2022}. However, significant contributions have been added to this article with respect to the original work, in order to improve our collaborative transportation framework and demonstrate its applicability in more detail. First of all, we introduce a novel human rotation intention algorithm based on body movements to enable complex co-transportation tasks that require the rotational motion of the object. Moreover, in order to explain our developed framework thoroughly, we performed new experiments by comparing it with the aforementioned existing approaches during co-transportation of the two extremities of the object deformability range (i.e, purely rigid aluminum rod and highly deformable rope). These experiments aim at showing that both types of feedback are essential for an effective co-transportation of objects with different deformability. Finally, we validated the potential of our framework with an additional experiment where a manikin on a deformable sheet was transported in collaboration with a mobile robot.

The remainder of this article is organized as follows: first, Section~\ref{sec:related_work} covers the background literature related to collaborative carrying. The details of the proposed control framework and the system overview are presented in Sections~\ref{sec:methodology} and~\ref{sec:experimental_setup}. Then, in Section~\ref{sec:preliminary_experiments}, our controller is demonstrated and compared to two existing approaches when co-transporting objects representing the deformability extremes, i.e., a purely rigid and a highly deformable. Later, Section~\ref{sec:experiments} reports the details of the conducted user study with a partially deformable object and the obtained quantitative and qualitative results. Finally, Section~\ref{sec:practical_showcase} describes an additional experiment where the applicability of our framework is shown and Section~\ref{sec:discussion} discusses the overall framework performance and draws the conclusions of the study. 

\section{RELATED WORK}
\label{sec:related_work}

This section presents a literature review about the previously mentioned challenges (i.e., load sharing, rotation-translation ambiguity and deformable object manipulation) in the context of collaborative transportation.

To address the load sharing problem, in~\cite{ikeura1995variable}, a variable impedance is used to control the robot's end-effector motion, where the damping switches between two discrete values based on end-effector velocity thresholds. Duchaine and Gosselin also adjust the virtual damping of the controller according to the human acceleration/deceleration intention which is predicted by using the time-derivative of the force at the robot's end-effector~\cite{Duchaine}. Then, the same authors extend their previous work by adding a stability observer that requires an online estimation of the human arm stiffness~\cite{duchaine2009safe}. In \cite{lecours2012variable}, mass and damper parameters of a variable admittance control are tuned online. Here, the issues related to using the time-derivative of the force as sensor for human intention are highlighted, and a new criterion for inferring the intended human motion is presented. 

\begin{figure*}
 \centering
    {\includegraphics[width=0.95\textwidth, trim=0cm 4.9cm 0cm 0cm, clip=true]{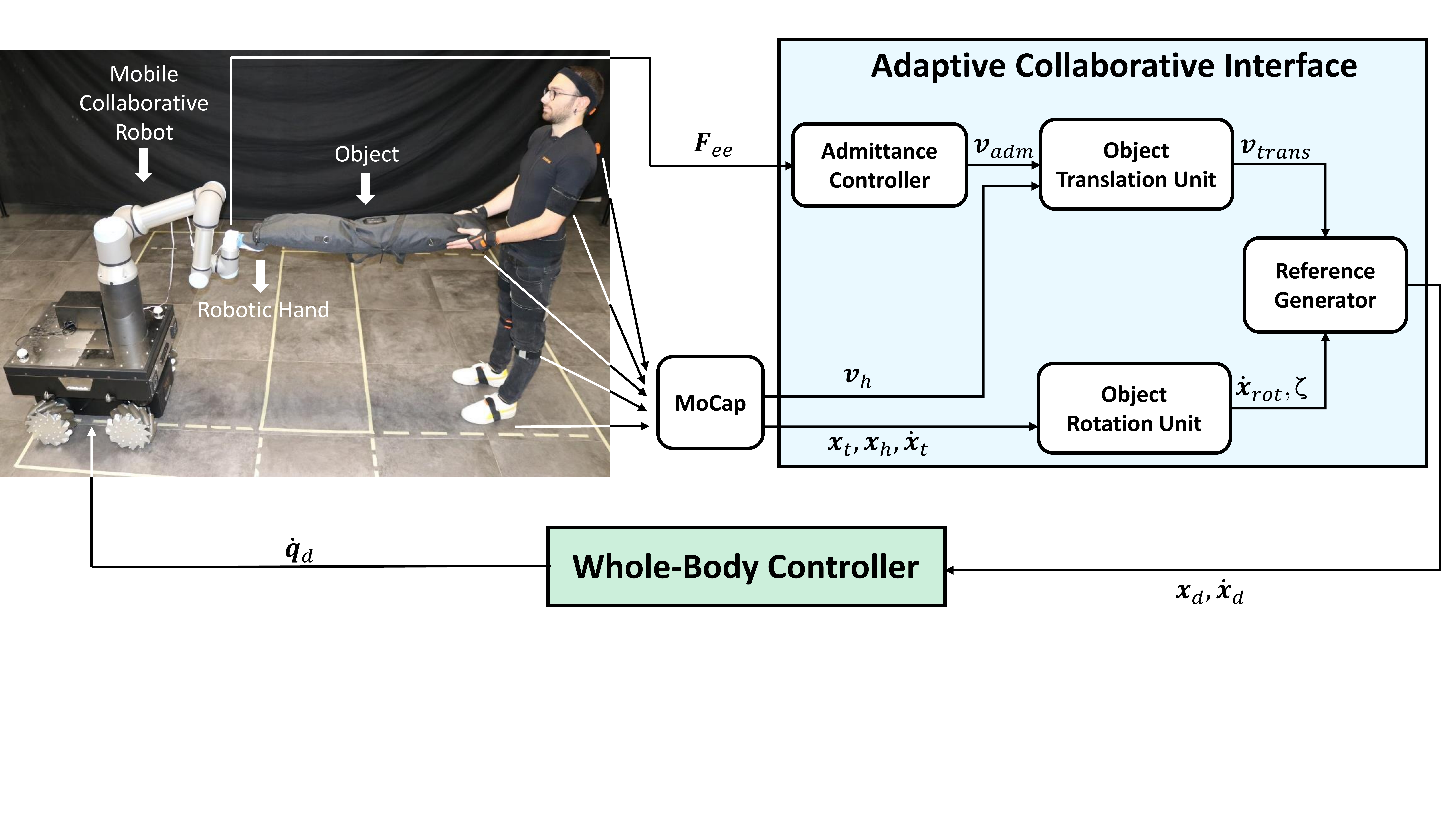}}
    \caption{The overall control architecture of the proposed framework.}
    \vspace{-0.0cm}
    \label{fig:system_overview_fig}
\end{figure*}

Although the works discussed thus far deal with the load sharing problem, they assume a follower behaviour of the robot, which limits the level of proactivity it can achieve. Conversely, an approach based on programming by demonstration (PbD) is used in~\cite{evrard2009teaching} to show follower and leader behaviours to a robot during a cooperative lifting task. The two behaviours are encoded through Gaussian Mixture Models (GMMs), and Gaussian Model Regression (GMR) is used to reproduce them during a user study. Bussy et al. utilize an admittance control law, where velocity and position error terms are used inside the admittance law to let the robot follow pre-defined motion primitives, which are triggered through heuristics based on velocity thresholds~\cite{bussy2012proactive}. Later, this proactive approach is combined with visual servoing in order to co-transport a rigid table while keeping a ball on top~\cite{agravante}. In this work, the vision-based controller is responsible to balance the ball by controlling only the height and the roll angle of the side that the robot carries. Similarly, Xinbo et al. present a hybrid control framework using visual and force sensing for collaborative carrying tasks of rigid objects~\cite{xinbo}. This proposed framework is designed to proactively follow the human operator by estimating the human motion intention. However, this approach requires the knowledge of the dynamics of the object being carried.

Regarding the translation/rotation ambiguity, there are two main approaches used to tackle this issue~\cite{Karayiannidis}: virtual constraints methods and techniques that consider different motion modes and switch between them. In~\cite{Karayiannidis}, the latter approach is used. The authors define a translation and a rotation mode, and they present two possible methods to switch between them. Instead, a virtual non-holonomic constraint is proposed by Takubo et al. in~\cite{Takubo2002}, consisting of a virtual wheel located at the robot's end-effector. By using this model, lateral movements are not allowed and the object is moved as if it was positioned on a cart with fixed passive wheels.

All the previously mentioned works consider just the rigid objects and only a few studies address the challenge of co-manipulating deformable objects in the literature~\cite{sanchez2018robotic}. Maeda et al. present an impedance control method by merging force and visual feedback where the desired position of the robot is estimated by a minimum jerk model of human hand motion~\cite{Maeda}. The proposed framework is validated through a co-transportation of a rubber pipe in a simplistic point-to-point transportation scenario. However, in a recent work~\cite{jerk_model}, it is revealed that minimum jerk model is not an appropriate fit for collaborative manipulation tasks. Similar to previous work, Kruse et al. propose a controller based on combination of force and visual feedback for the co-transportation of a cloth~\cite{kruse2015collaborative}. When the cloth is taut, the force-feedback allows robot compliance. In contrast, when the object is slack due to human movements, the visual-feedback enables the generation of motions that help the robot recover tautness of the cloth. Nonetheless, the applicability of this approach is limited to a cloth. In another study~\cite{delpreto2019sharing}, DelPreto and Rus design a control framework based on EMG information of the upper arm to assist the human operator in a collaborative lifting task. Despite the promising results obtained during the co-lifting task, adapting this method to co-transportation tasks involving 3D movements is not possible.

\section{METHODOLOGY}
\label{sec:methodology}

The overall control architecture of the presented system (see Fig.~\ref{fig:system_overview_fig}) is composed of two main modules: (1) a Whole-Body Controller which generates desired joint velocities for the mobile base robotic platform and (2) an Adaptive Collaborative Interface that calculates reference inputs for the whole-body controller based on the perceived sensory information. These two components and the interaction between them are explained in detail in the following subsections.

\subsection{Whole-Body Controller}

The Whole-Body Controller employed on the robot is based on the solution of a Hierarchical Quadratic Program (HQP) composed of two tasks. The formulation of the problem as a HQP allows to exploit the redundancy of the robot, since it has $m=n_b+n_a>6$ DoFs. The cost function of the higher priority task is written as \cite{seraji1990improved} (dependencies are dropped):
\begin{align}
    \mathcal{L}_1 = || \dot{\boldsymbol{x}_d} + \boldsymbol{K}({\boldsymbol{x}_d} -{\boldsymbol{x}}) - \boldsymbol{J}\dot{\boldsymbol{q}}||^2_{\boldsymbol{W}_{1}} + ||k\dot{\boldsymbol{q}}||^2_{\boldsymbol{W}_{2}},
\end{align}
where the vector of the whole-body joint velocities $\boldsymbol{\dot{q}}$ $\in$ $\mathbb{R}^{m}$ is the optimization variable, ${\boldsymbol{J}}$ $\in$ $\mathbb{R}^{6\times m}$ is the whole-body Jacobian, ${\boldsymbol{x}}$ $\in$ $\mathbb{R}^{6}$ is the current end-effector pose, ${\boldsymbol{W}_1}$ $\in$ $\mathbb{R}^{6\times6}$, ${\boldsymbol{W}_2}$ $\in$ $\mathbb{R}^{m\times m}$, and ${\boldsymbol{K}}$ $\in$ $\mathbb{R}^{6\times6}$ are diagonal positive definite matrices and $k$ $\in$ $\mathbb{R}_{>0}$ is the so-called damping factor \cite{chiaverini1992weighted}, which depends on the manipulability index of the arm \cite{deo1995overview, hollerbach1987redundancy, wampler1986manipulator}. Then, the joint velocities of the secondary task are computed as the negative gradient w.r.t. $\boldsymbol{q}$ of (\ref{eq:secondary_task})~\cite{nakanishi2005comparative, wu2021unified}, where ${\boldsymbol{W}_3}$ $\in$ $\mathbb{R}^{m\times m}$ is a diagonal positive semidefinite matrix and ${\boldsymbol{q}_{def}}$ is a default joint configuration: 
\begin{equation}
\label{eq:secondary_task}
    ||\boldsymbol{q}_{def} - \boldsymbol{q}||^2_{\boldsymbol{W}_{3}}.
\end{equation}
These velocities are later projected in the null-space of the primary task.

This formulation allows to obtain the desired whole-body joint velocities $\boldsymbol{\dot{q}}_d$ $\in$ $\mathbb{R}^{m}$ that distribute the movements between base and arm. The primary task guarantees the tracking of the desired end-effector motion ($\boldsymbol{\dot{x}}_d$ and ${\boldsymbol{x}_d}$). On the other hand, the secondary task keeps the arm close to the joints configuration $\boldsymbol{q}_{def}$, which can be chosen in order to ensure a task-related robot posture during the co-carry.

\subsection{Adaptive Collaborative Interface}

The detailed architecture of our Adaptive Collaborative Interface (ACI) that handles objects with different deformation properties is illustrated in Fig.~\ref{fig:system_overview_fig}. The presented interface can be subdivided into four operational units, namely an \emph{Admittance Controller}, which computes a reference translational velocity based on the force transferred through the object, an \emph{Object Translation unit} that merges the velocity calculated by the Admittance Controller together with the reference velocity produced by the MoCap system, an \emph{Object Rotation unit}, which finds a reference twist based on the human torso and hand movements, and a \emph{Reference Generator} that sends the actual reference pose and twist to the Whole-Body Controller based on the input coming from both Object Translation and Rotation units.

\subsubsection{Admittance Controller}

In this work, a standard admittance controller is utilized. It implements the following transfer function expressed in Laplace domain:
\begin{equation}
  \boldsymbol{{V}}_{adm}(s) = \frac{\boldsymbol{F}_H(s)}{\boldsymbol{M}_{adm}s+\boldsymbol{D}_{adm}},
\end{equation}

\noindent where $\boldsymbol{M}_{adm}$ and $\boldsymbol{D}_{adm}$  $\in$ $\mathbb{R}^{3\times3}$ are the desired mass and damping matrices, $s$ is the Laplace variable, $\boldsymbol{F}_H(s) \in \mathbb{R}^{3}$ is the Laplace transform of the measured forces and $\boldsymbol{V}_{adm}(s) \in \mathbb{R}^{3}$ is the Laplace transform of the admittance reference translational velocity.

\subsubsection{Object Translation Unit}

As mentioned earlier, the main contribution of this work is to enable human-robot co-transportation of objects having different deformability. Due to such property of the object, relying just on haptic information may not be sufficient for an effective coordination of the co-carrying task. To address this issue, we introduce the \emph{Object Translation unit}, which computes a translational velocity reference ($\boldsymbol{{v}}_{trans}$) based on the admittance reference velocity ($\boldsymbol{{v}}_{adm}$) and the human hand velocity ($\boldsymbol{{v}}_{h}$) measured by the MoCap system. This unit calculates an adaptive index during the collaboration as follows:
\begin{align}
\label{eq:alpha_calculation}
    \alpha =1-\frac{||\int_{t_{c}-W_{l}}^{t_{c}} \boldsymbol{{v}}_{adm}(t) \,dt||\ }{||\int_{t_{c}-W_{l}}^{t_{c}} \boldsymbol{{v}}_{h}(t) \,dt||\  + \epsilon} ,
\end{align}
\noindent where $\alpha \in [0,1]$ is the adaptive index, $t_c$ is the current time, $W_l$ is the length of the sliding time window and $\epsilon$ is a small number used to avoid the problem of division by zero. This index allows to understand whether the object is non-deformable ($\alpha = 0$), deformable ($\alpha = 1$) or partially deformable ($\alpha \in (0,1)$). Note that,  $\alpha$ is  saturated at 0.

Subsequently, $\alpha$ is used on-the-fly to regulate the contribution of $\boldsymbol{{v}}_{h}$ to $\boldsymbol{{v}}_{trans}$ using:
\begin{align}
\label{eq:velocity_calculation}
    \boldsymbol{{v}}_{trans} = \boldsymbol{{v}}_{adm} + \alpha\boldsymbol{{v}}_{h}.
\end{align}

This formulation allows to benefit from two different sources of information. 
In order to clarify how the \textit{Object Translation unit} deals with objects having different deformation properties, it can be informative to analyze three distinct cases. If the object being carried does not transmit any force applied by the human (e.g. a loose rope), $\alpha$ is closer to 1.  In that case, $\boldsymbol{{v}}_{adm}$ is always close to zero and consequently, $\boldsymbol{{v}}_{trans} \approx \boldsymbol{{v}}_{h}$. On the other hand, if the carried object is capable of transferring the forces applied by the human (e.g., a rigid box), human and robot will move rigidly together ($\boldsymbol{{v}}_{h} \approx \boldsymbol{{v}}_{adm}$). In this situation, $\alpha$ is closer to 0, resulting in $\boldsymbol{{v}}_{trans} \approx \boldsymbol{{v}}_{adm}$. The two examples presented so far are the extreme cases of highly deformable and purely rigid objects, respectively. When the objects exhibit intermediate characteristics in terms of deformability, $\alpha$ will take values between 0 and 1, resulting in a reference translational velocity computed by a combination of $\boldsymbol{{v}}_{adm}$ and $\boldsymbol{{v}}_{h}$.

\subsubsection{Object Rotation Unit}

In this work, human torso and hand movements are utilized to detect the intention of rotating an object. Specifically, to initiate a rotational movement, the human agent should turn his/her torso towards a desired rotation angle. At this point, this unit plays a key role for both detecting the human rotation intention and planning the trajectory to the intended pose.

\renewcommand{\algorithmicrequire}{\textbf{Input:}}
\renewcommand{\algorithmicensure}{\textbf{Output:}}

\begin{algorithm}
\caption{Human Rotation Intention Algorithm}\label{alg:cap}
\begin{algorithmic}
\Require $\theta_{h}^{t}, \theta_{h}^{w}, \theta_{t}^{w}, \dot{\theta_{t}^{w}}, \boldsymbol{T}_{t}^{w}$
\Ensure $\zeta, \boldsymbol{T}_{t,det}^{w}$
\State $Initialization:$
\State $\theta_{h,l} \gets 0, \theta_{h,u} \gets 0, \theta_{t,l} \gets 0, \theta_{t,u}
\gets 0$
\State $Control$ $loop:$

\If{$|\theta_{h}^{t}|$ $>$ $lower$ $angle$ $threshold$}
    \State $\theta_{h,l} \gets \theta_{h}^{w},\theta_{t,l} \gets \theta_{t}^{w}$
    \While {$|\theta_{h}^{t}|$ $>$ $lower$ $angle$ $threshold$}
        \State $\theta_{h,u} \gets \theta_{h}^{w},\theta_{t,u} \gets \theta_{t}^{w} $
        \State $\Delta_{\theta_{h}} \gets |\theta_{h,u}-\theta_{h,l}|$
        \State $\Delta_{\theta_{t}} \gets |\theta_{t,u}-\theta_{t,l}|$
        \If{$|\theta_{h}^{t}|$ $>$ $upper$ $angle$ $threshold$ and $\Delta_{\theta_{t}}$ $>$ $\Delta_{\theta_{h}}$ and             $|\dot{\theta_{t}^{w}}|$ $<$ $velocity$ $threshold$}
            \State $\zeta \gets 1$ \Comment{Human rotation intention}
            \State $\boldsymbol{T}_{t,det}^{w} \gets \boldsymbol{T}_{t}^{w}$
        \Else
            \State $\zeta \gets 0$
        \EndIf
    \EndWhile
\Else
        \State $\zeta \gets 0$
        
\EndIf
\end{algorithmic}
\end{algorithm}

In order to address the rotation intention problem, Alg.~\ref{alg:cap} is developed in this work. Its main objective is to detect whether or not the human intends to rotate the object, which is represented through the boolean variable $\zeta$. Then, if the rotation intention is detected, the pose of the human torso at that moment ($\boldsymbol{T}^w_{t,det}$) is recorded by the algorithm. The relative yaw angle between the human torso and the hand ($\theta_{h}^{t}$) can be considered as primary source of information in our algorithm. However, employing it alone is not sufficient to infer the source of its change. For instance, it can be altered by only rotating the hand or by rotating the torso unintentionally due to natural body movements, which might happen during co-carrying tasks. To overcome this issue, our algorithm also uses torso and hand yaw angles expressed in world frame ($\theta_{h}^{w}$ and $\theta_{t}^{w}$ respectively) for identifying the one responsible of a change in $\theta_{h}^{t}$. In addition, the torso yaw velocity ($\dot{\theta_{t}^{w}}$) is utilized to infer whether or not the torso movement is ended.

\begin{figure}
    \centering
    \resizebox{1.0\columnwidth}{!}{\rotatebox{0}{\includegraphics[trim=9cm 9cm 10cm 0cm, clip=true]{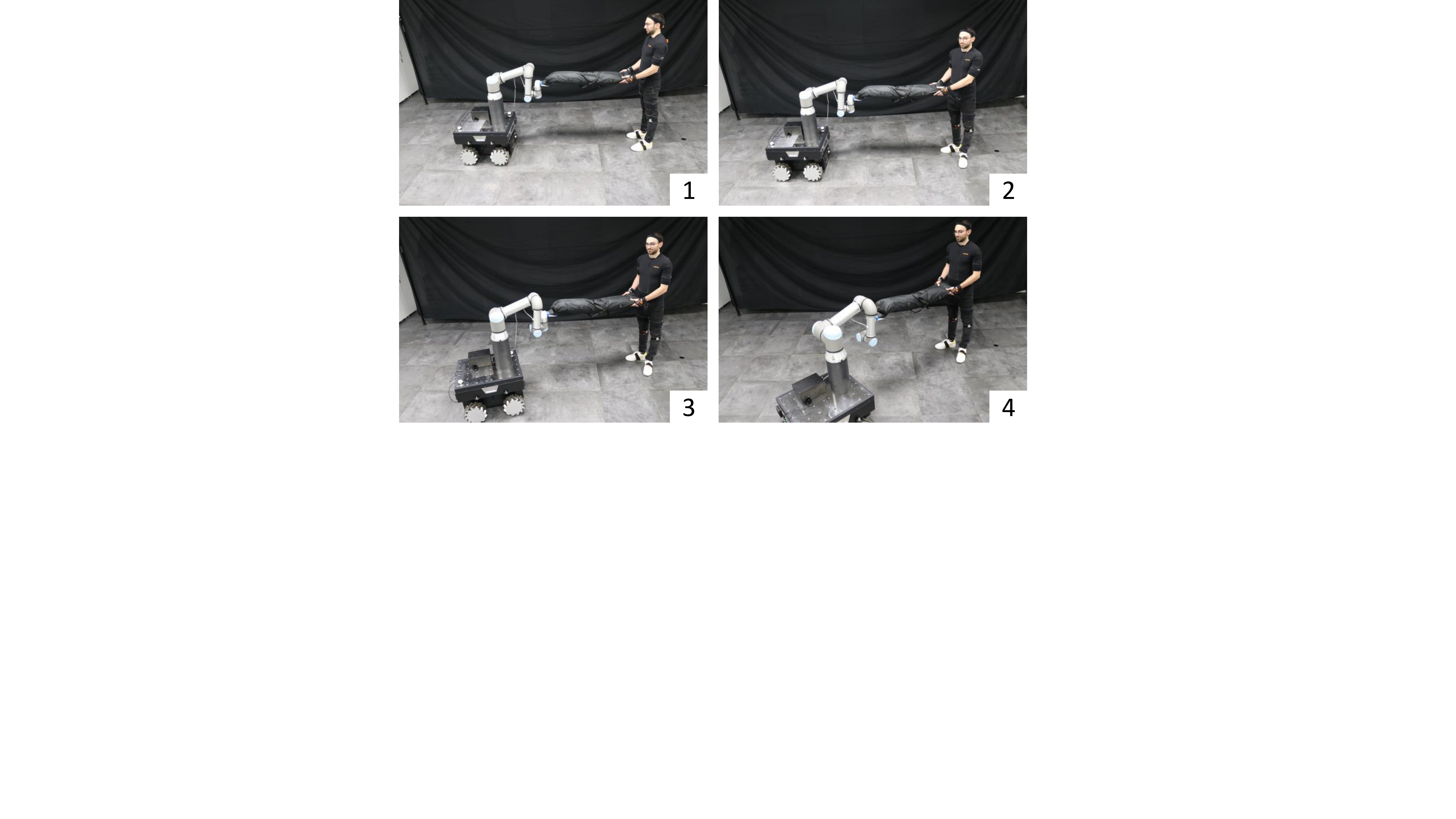}}}
    \caption{Snapshots of the rotation motion. The human operator carries the object collaboratively without rotation intention (1), then he intends to rotate by turning his torso to the desired rotation angle and this intention is detected by our algorithm (2). Next, the robot starts to move along its planned trajectory (3), and the robot reaches the desired pose (4).}
    \label{fig:rotation_fig}
    \vspace{-0.0cm}
\end{figure}

If human rotation intention is detected by the algorithm, \emph{Object Rotation unit} starts to plan the trajectory to the desired position and orientation as illustrated in Fig.~\ref{fig:rotation_fig}. The desired robot's end-effector transformation w.r.t. the world frame can be computed as
\begin{align}
    \boldsymbol{T}_{r,des}^{w} = \boldsymbol{T}_{t,det}^{w}\boldsymbol{T}_{r,i}^{t,i},
\end{align}
where $\boldsymbol{T}_{r,i}^{t,i}$ is the initial transformation between the robot's end-effector and the human torso. With this derivation, the initial configuration between robot's end-effector and human torso is  preserved, when a rotation intention is detected. Indeed, it is assumed that the initial configuration is the desirable one for the human during collaboration. Finally, a point-to-point motion trajectory planning is implemented by means of a third-order polynomial to have smoother trajectories between the actual robot pose and the desired robot pose. 

\subsubsection{Reference Generator}
This unit is in charge of sending the actual reference pose and twist to the Whole-Body Controller, based on the outcome of the human rotation intention detection. They are computed as follows:
\begin{align}
    \boldsymbol{\dot{x}}_{d} =  \zeta\boldsymbol{\dot{x}}_{rot} + (1-\zeta)\boldsymbol{\dot{x}}_{trans},
\end{align}
\begin{align}
    \boldsymbol{{x}}_{d}= \int_{0}^{t_{c}} \boldsymbol{\dot{x}}_{d}(t) \,dt\ ,
\end{align}
\noindent where $\boldsymbol{\dot{x}}_{trans} = [\boldsymbol{v}_{trans}^{T}, \boldsymbol{0}^{T}]^T$.

\section{SYSTEM OVERVIEW}
\label{sec:experimental_setup}

In this work, the robotic platform Kairos (see Fig.~\ref{fig:system_overview_fig}) was used for our experiments. It consists of an Omni-directional Robotnik SUMMIT-XL STEEL mobile base, and a high-payload (16 kg) 6-DoFs Universal Robot UR16e arm attached on top of the base.

In order to track human movements during co-carry, we decided to use the Xsens as a MoCap system (see Fig.~\ref{fig:system_overview_fig}). Though our framework can operate with different MoCap systems, such as vision-based, Xsens was preferred in this study due to its precision. It is composed of seventeen Inertial Measurement Units (IMUs) placed on specific parts of the human body. Thanks to this system, it is possible to get real-time measurements of the human hand and torso twists, that are fed to our controller as described in Section~\ref{sec:methodology}.

\section{EXTREMITIES}
\label{sec:preliminary_experiments}

\begin{figure*}[h]
    \centering
    \resizebox{1\textwidth}{!}{\rotatebox{0}{\includegraphics[trim=1.8cm 9.5cm 1.8cm 9cm, clip=true]{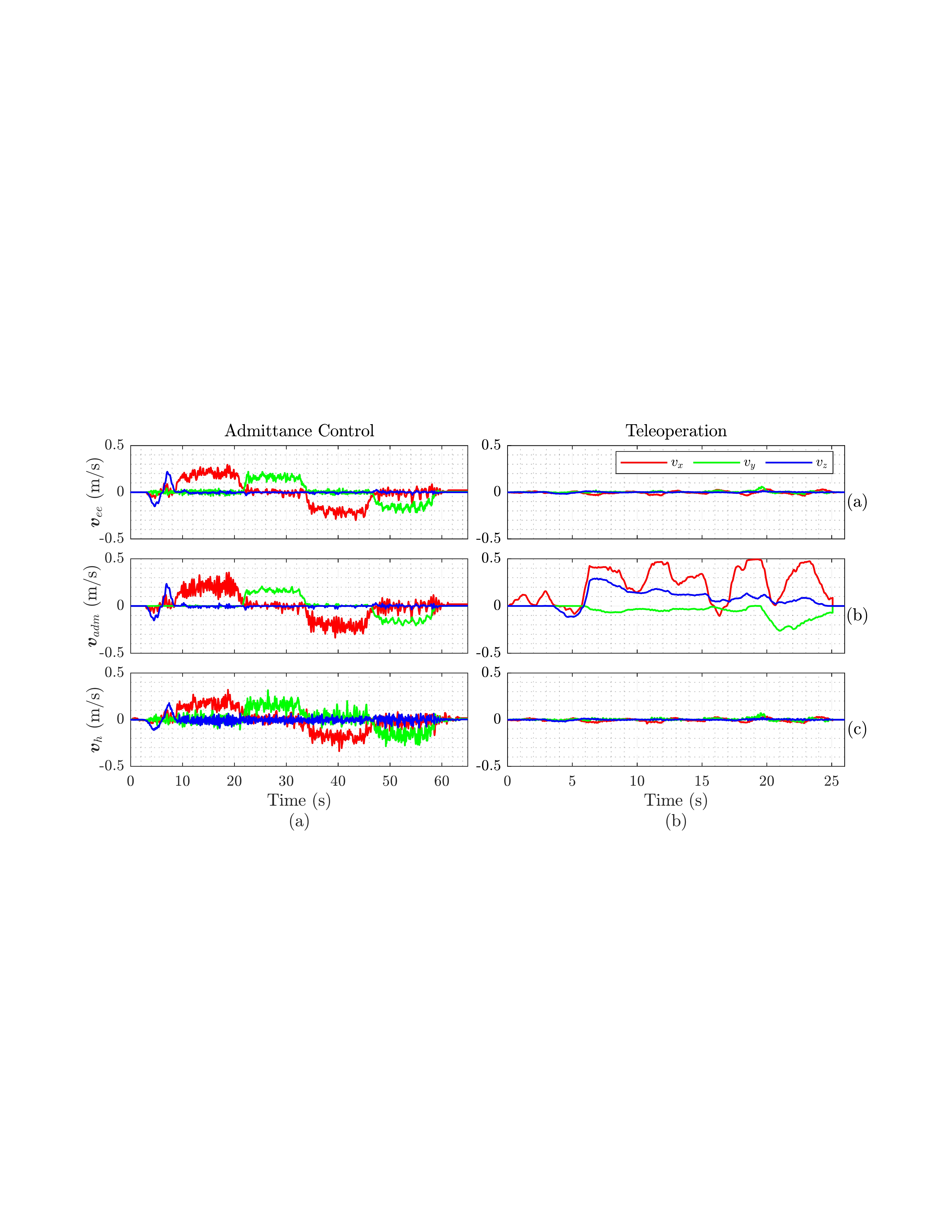}}}
    \caption{Results of the co-transportation experiments of a rigid aluminum profile with an admittance control and teleoperation framework. The graphs show (a) the end-effector velocity $\boldsymbol{{v}}_{ee}$, (b) the admittance reference velocity $\boldsymbol{{v}}_{adm}$, and (c) the human hand velocity $\boldsymbol{{v}}_{adm}$.}
    \label{fig:rigid_adm_tel}
    \vspace{-0.0cm}
\end{figure*}  

In this article, we consider the two extremities of the object deformability range as \textit{purely rigid}, that transmits the applied wrenches from one end to the other end of the object without any loss, and \textit{highly deformable}, which does not transfer any haptic information. The use of haptic-based controllers or teleoperation-like frameworks (where human movements are mirrored and replicated by the robot) alone are not sufficient to perform a co-transportation task for both the extreme cases. While the haptic-based controllers are suitable for rigid objects, they are not competent during co-transportation of deformable objects due to the lack of haptic cues. On the other hand, teleoperation-like frameworks enable to perform co-transportation of highly deformable objects without requiring haptic information. However, they fail in the other extreme condition because the rigid connection between the human and the robot does not allow them to initiate the motion. In the following subsections, we demonstrate the discussed performances of haptic-based controllers and teleoperation frameworks in these two extremities, and additionally, we illustrate how the proposed controller can handle both thanks to its adaptive nature. In the experiments performed in this section, the human partner is simply asked to carry out co-transportations with the robot along the three directions for fixed distances.

\subsection{Purely Rigid Object}

In these experiments, we evaluated the performances of 3 different frameworks, namely the admittance controller, the teleoperation\footnote{As mentioned erlier, in this article, teleoperation refers to a control architecture through which human movements are mirrored and replicated by the robot during co-carrying.}, and the proposed controller (ACI) during the co-transportation of an aluminum rod that was rigidly connected to the end-effector of the robot. We used $\boldsymbol{{v}}_{adm} = \boldsymbol{{v}}_{trans}$ for the admittance controller and $\boldsymbol{{v}}_{h} = \boldsymbol{{v}}_{trans}$ in the case of teleoperation, instead of combining $\boldsymbol{{v}}_{adm}$ and $\boldsymbol{{v}}_{h}$ by using the adaptive index (see Eq.~\ref{eq:velocity_calculation}) as in our proposed controller. 

\begin{figure}
    \centering
    \resizebox{1.0\columnwidth}{!}{\rotatebox{0}{\includegraphics[trim=0cm 0cm 0cm 0cm, clip=true]{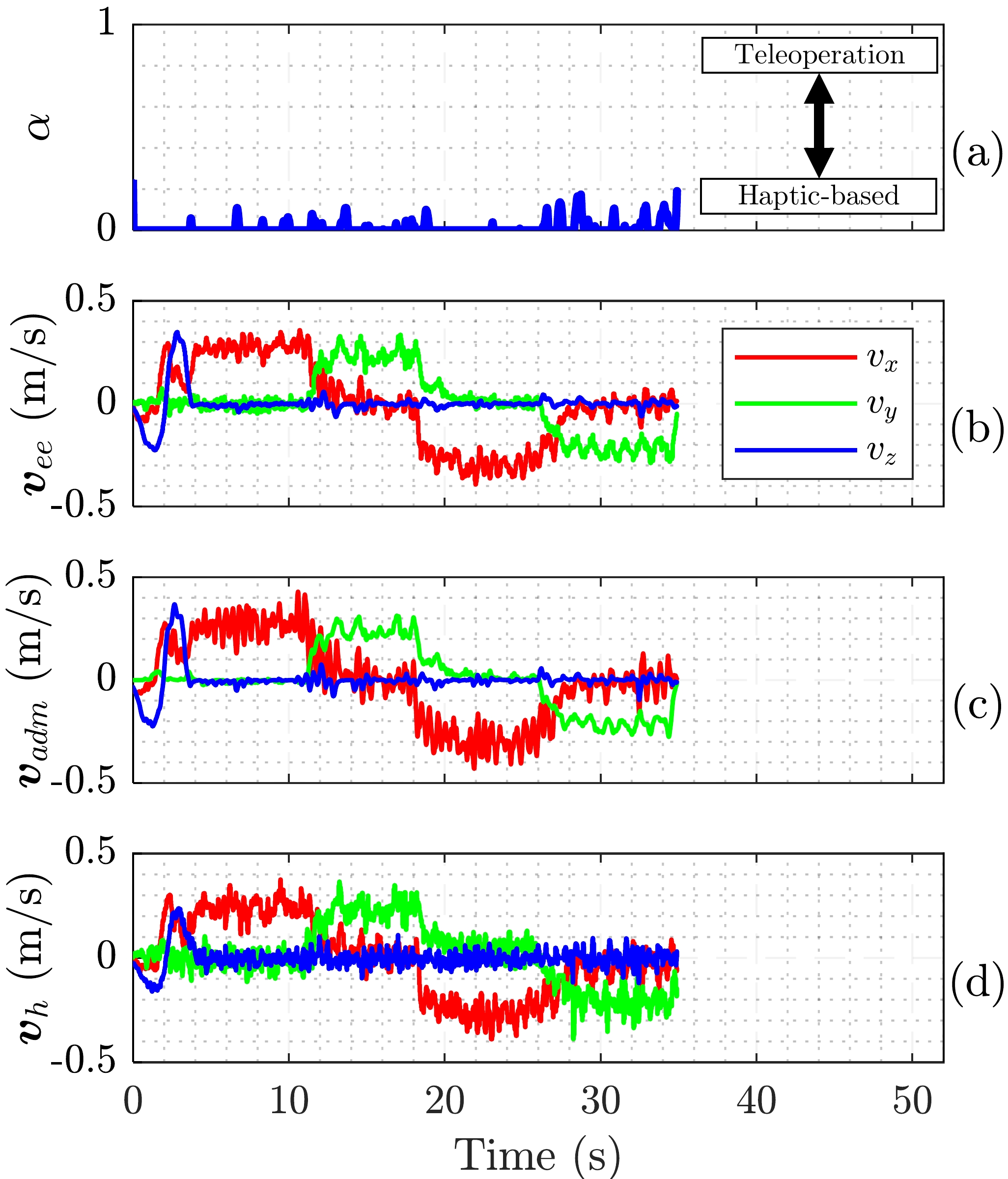}}}
    \caption{(a) The adaptive index $\alpha$, (b) the end-effector velocity $\boldsymbol{{v}}_{ee}$, (c) the admittance reference velocity $\boldsymbol{{v}}_{adm}$, and (d) the human hand velocity $\boldsymbol{{v}}_{h}$ during co-transportation of a rigid aluminum profile with the proposed framework.}
    \label{fig:rigid_aci}
    \vspace{-0.0cm}
\end{figure}

\begin{figure*}
    \centering
    \resizebox{1\textwidth}{!}{\rotatebox{0}{\includegraphics[trim=1.8cm 9.5cm 1.8cm 9cm, clip=true]{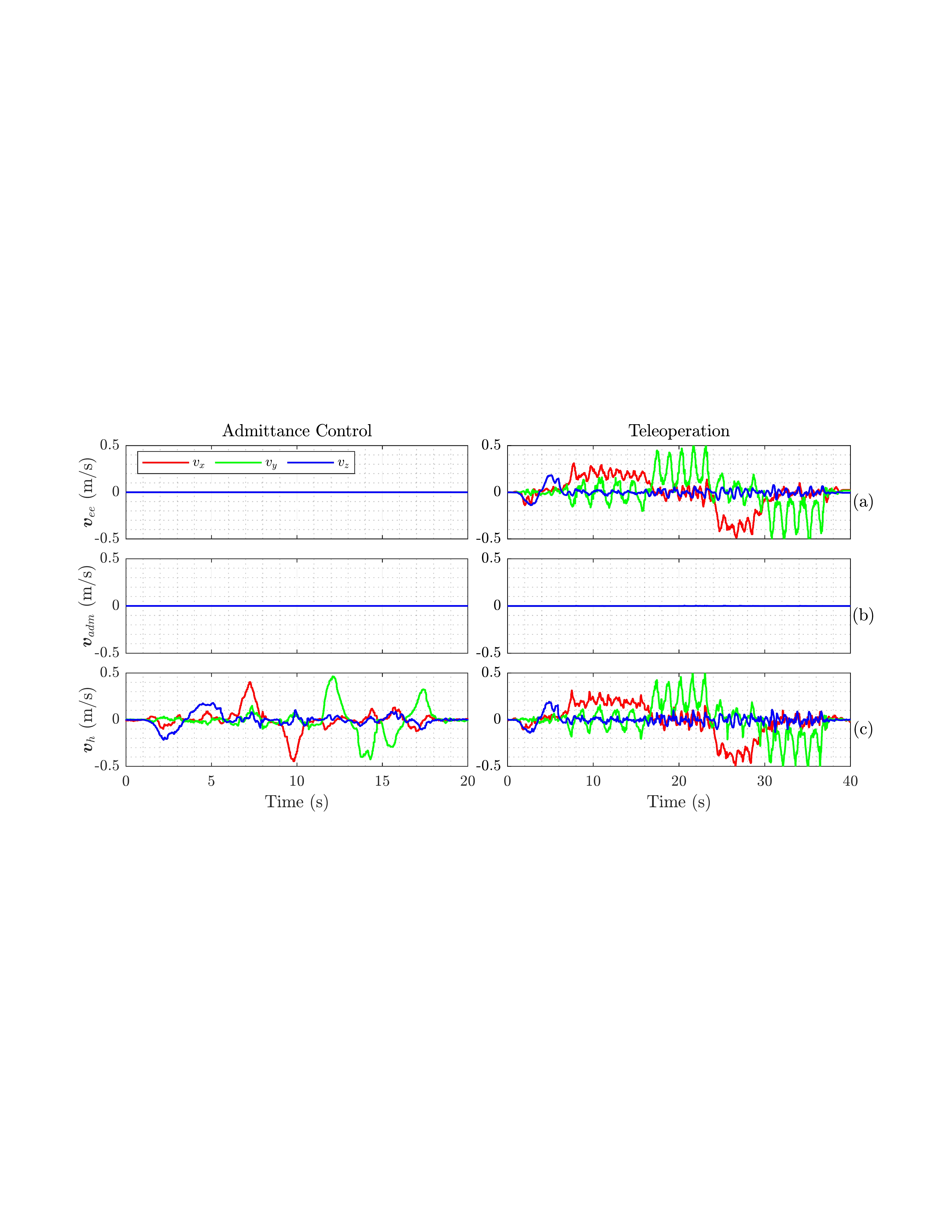}}}
    \caption{Results of the co-transportation experiments of a highly deformable rope with an admittance control and teleoperation framework. The graphs show (a) the end-effector velocity $\boldsymbol{{v}}_{ee}$, (b) the admittance reference velocity $\boldsymbol{{v}}_{adm}$, and (c) the human hand velocity $\boldsymbol{{v}}_{adm}$.}
    \label{fig:deformable_adm_tel}
    \vspace{-0.0cm}
\end{figure*}

Fig.~\ref{fig:rigid_adm_tel} shows the results obtained from the co-carry experiments of the aluminum rod with the admittance controller and teleoperation only. In the case of teleoperation, the robot stood still and the task could not be executed due to the rigid connection between the human and the robot. Indeed, this connection 
prevented the human from generating $\boldsymbol{{v}}_{h}$, even if he applied forces to the object (see Fig.~\ref{fig:rigid_adm_tel}b,~\ref{fig:rigid_adm_tel}c). On the other hand, when the admittance controller was employed, the co-transportation task was performed successfully in all the directions. In addition, it can also be realized that $\boldsymbol{{v}}_{ee}$ was almost equal to $\boldsymbol{{v}}_{h}$ in both trials, since the human hand and the end-effector were connected rigidly to each other through the carried object.

The logged velocities and the adaptive index ($\alpha$) data during the co-carry of the same object with the proposed controller are presented in Fig.~\ref{fig:rigid_aci}. As in the admittance controller case, the human operator was able to carry the object in all the directions. As explained in Sec.~\ref{sec:methodology}, $\alpha$ changes on-the-fly according to the deformability of the object being carried. Here, it was computed almost equal to zero during the whole task, as anticipated, since $\boldsymbol{{v}}_{adm}$ and $\boldsymbol{{v}}_{h}$ were approximately equal to each other during the whole co-transportation task. The effect of $\boldsymbol{{v}}_{h}$ becomes negligible on the calculated reference velocity and the proposed controller behavior approaches the admittance controller to enable performing the task.

\subsection{Highly Deformable Object}

For the co-transportation experiments of the highly deformable object, we used an unstretched rope in order to avoid the transmission of wrenches applied by the human to the end-effector of the robot. Fig.~\ref{fig:deformable_adm_tel} depicts the recorded results during the experiments with the admittance controller and the teleoperation framework. The rope did not provide any haptic feedback to the robot during the motion, as can be seen from Fig.~\ref{fig:deformable_adm_tel}b. Therefore, the robot was not able to move with the admittance controller, whereas the co-transportation task was successfully performed with the teleoperation framework which uses the human hand motion as the source of information for the reference velocity to follow.

\begin{figure}
    \centering
    \resizebox{1.0\columnwidth}{!}{\rotatebox{0}{\includegraphics[trim=0cm 0cm 0cm 0cm, clip=true]{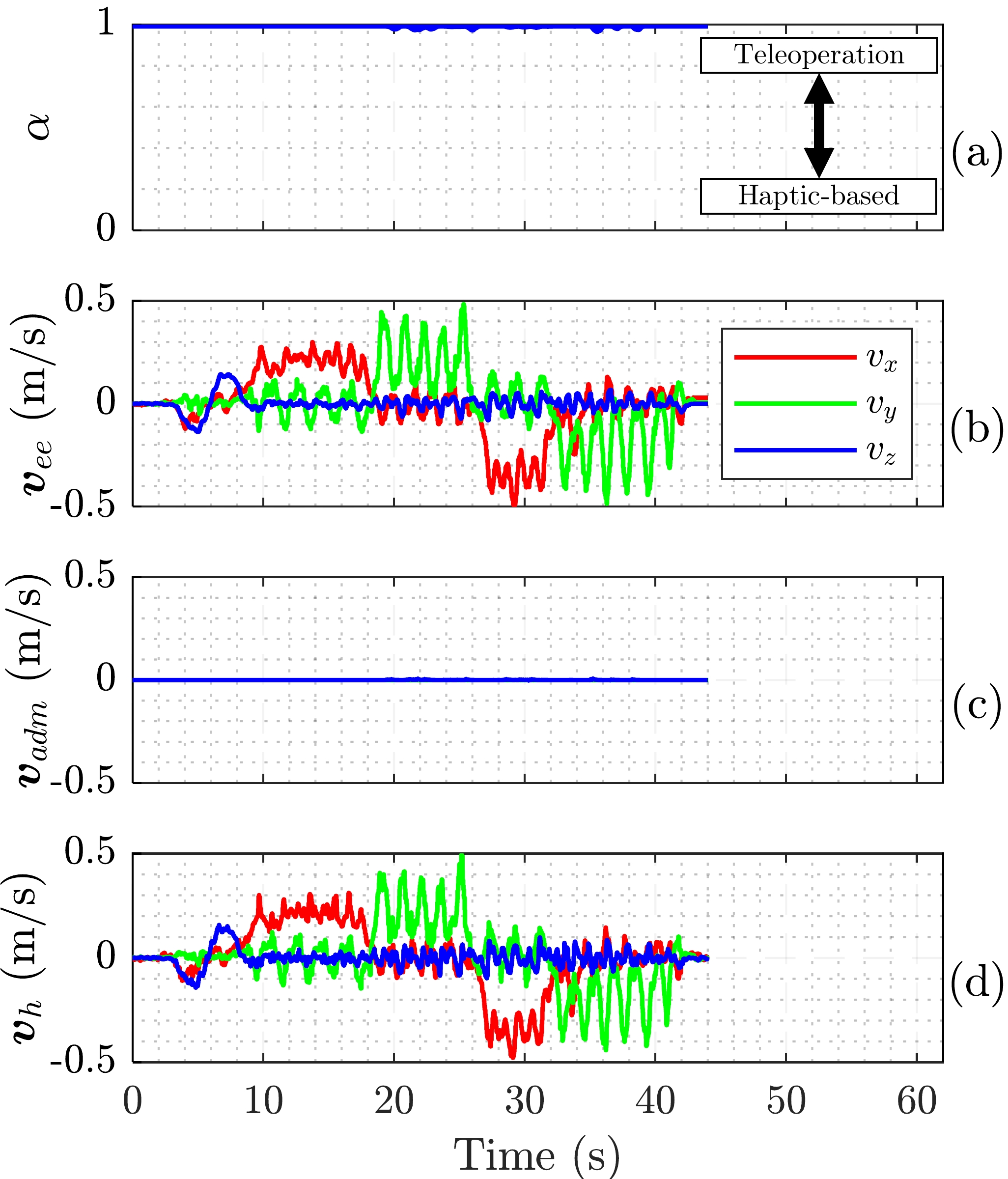}}}
    \caption{(a) The adaptive index $\alpha$, (b) the end-effector velocity $\boldsymbol{{v}}_{ee}$, (c) the admittance reference velocity $\boldsymbol{{v}}_{adm}$, and (d) the human hand velocity $\boldsymbol{{v}}_{h}$ during co-transportation of a highly deformable rope with the proposed framework.}
    \label{fig:deformable_aci}
    \vspace{-0.0cm}
\end{figure}  

The data obtained from the co-carry of the same rope with our controller is exhibited in Fig.~\ref{fig:deformable_aci}. Unlike the rigid case, $\alpha$ was computed as almost always equal to 1 on this opposite side of the object deformability range ($\boldsymbol{{v}}_{adm} \ll \boldsymbol{{v}}_{h}$ in Eq.~\ref{eq:alpha_calculation}). This adaptive feature of the controller allowed to handle this extremity by adding $\boldsymbol{{v}}_{h}$ based on $\alpha$ into $\boldsymbol{{v}}_{trans}$.

\section{USER STUDY: PARTIALLY DEFORMABLE OBJECT}
\label{sec:experiments}

To verify the effectiveness of our proposed framework, we investigated the performance of our Adaptive Collaborative Interface (ACI) during a human-robot co-carrying task of a partially deformable object. For this experiment, a bag (44 cm length, 9.5 cm width, 5 cm height) filled with styrofoam packing peanuts was used (see Fig.~\ref{fig:system_overview_fig}). It features an approximately rigid behaviour if stretched along its larger dimension (when it is pulled), while it is rather partially deformable if stressed along the other directions. Because the co-transportation of the rigid objects cannot be performed with teleoperation frameworks as demonstrated in the previous section, we used an admittance controller as a baseline for our experiments.

\begin{figure}[t!]
    \centering
    \resizebox{1\columnwidth}{!}{\rotatebox{0}{\includegraphics[trim=3cm 0.5cm 4cm 3cm, clip=true]{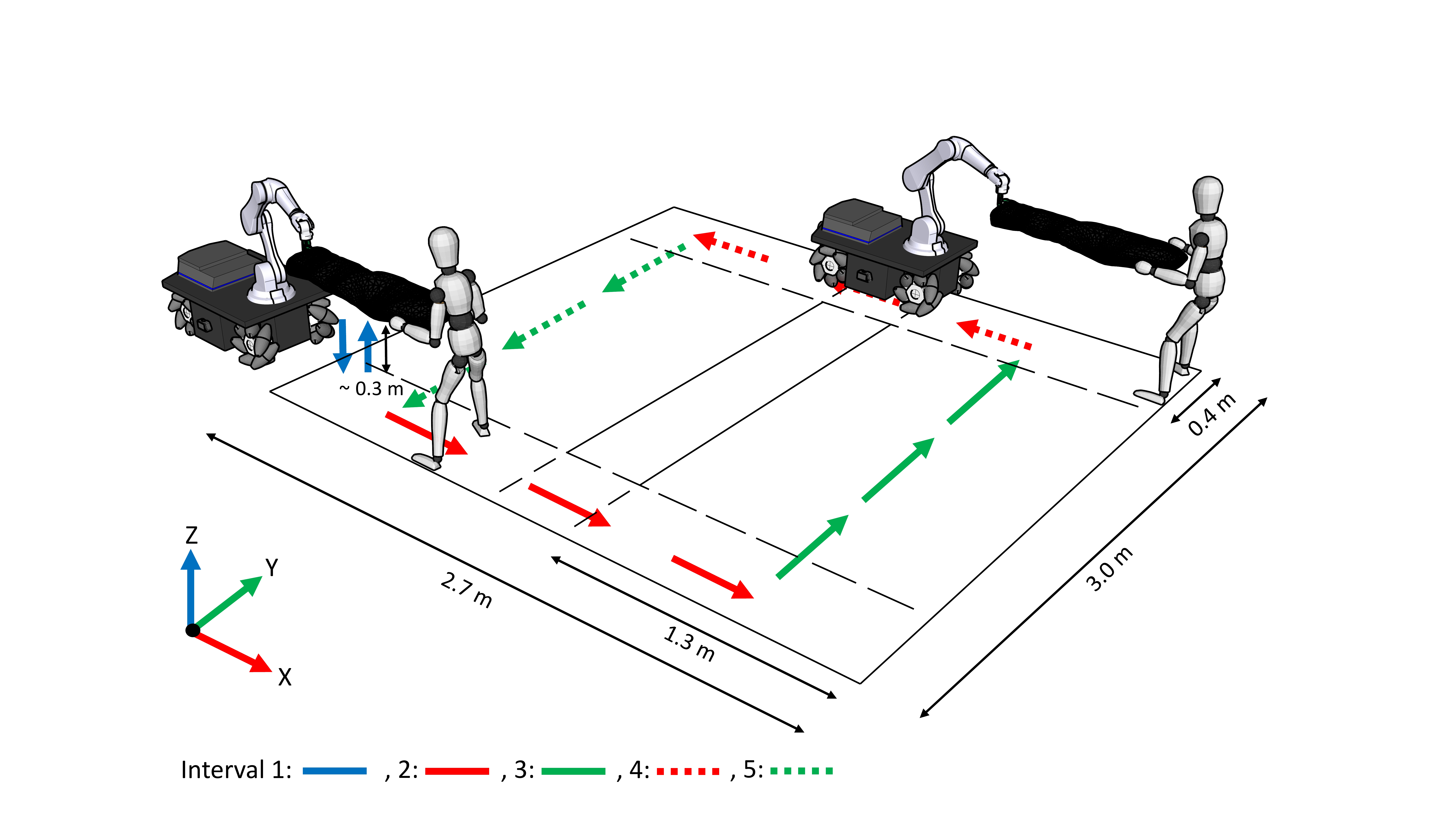}}}
    \caption{Illustration of the experiment and its intervals, in which an object is co-transported with a mobile manipulator along a path to be followed.}
    \vspace{-0.0cm}
    \label{fig:trajectory_and_object}
\end{figure}

\subsection{Experimental Procedure}

In the experiment, the participants were asked to co-carry a partially deformable object along a designed path in collaboration with the robot. An underactuated Pisa/IIT SoftHand was mounted at the end-effector of the robotic arm to grasp the object. The path comprises six main sub-movements: an initial down-up movement (2 of 6) followed by a square trajectory on a plane (4 of 6) parallel to the floor at a height that is customized based on the preference of each subject, as shown in Fig.~\ref{fig:trajectory_and_object}. The following guidelines were given to the participants before the beginning of the experiment:

\begin{itemize}
    \item The object must be kept inside the drawn path.
    \item If the object is tilted at the end of each sub-movement, the participant cannot start the following one. In this case, the participant must transmit a force to the robot by deforming the object to align the robot to the intended position and then start the following sub-movement.
    \item The task must be completed as fast as possible.
\end{itemize}

Each participant performed a total of 10 trials (2 controllers $\times$ 5 repetitions). The order of the trials was randomized not to affect the results with the learning effect. Before the experiment, a familiarization phase for both controllers was conducted until the subjects felt at their ease with the system. To guarantee human safety, mobile base velocities and the arm forces were constantly monitored and limited.

\subsection{Participants}

Twelve healthy volunteers, six males and six females, (age: $26.8 \pm 6.7$ years; mass: $64.9 \pm 16.2$ kg; height: $172.4 \pm 10.1$ cm)\footnote{Subject data is reported as: mean $\pm$ standard deviation.} were recruited for the experiments. After explaining the experimental procedure, written informed consent was obtained, and a numerical ID was assigned to anonymize the data. The whole experimental activity was carried out at the Human-Robot Interfaces and Physical Interaction (HRII) Lab, Istituto Italiano di Tecnologia (IIT), in accordance with the Declaration of Helsinki. The protocol was approved by the ethics committee Azienda Sanitaria Locale (ASL) Genovese N.3 (Protocol IIT\_HRII\_ERGOLEAN 156/2020).

\subsection{Controller Parameters}

In order to have a fair comparison between the ACI and the standard admittance controller, the same desired mass ($\boldsymbol{M}_{adm} = diag \{6,6,6\}$) and damping ($\boldsymbol{D}_{adm} = diag \{30,30,30\}$) were assigned for both controllers in this study. These parameters were selected based on a heuristic, in order to balance a trade-off between transparency and stability during the co-transportation.

Furthermore, the length of the sliding time window ($W_l$) which was used for the calculation of the adaptive index ($\alpha$) was selected as 0.25 s. This time length was tuned in order to compromise the delay in the detection of a change in the object deformability with its accurate identification. 

Regarding the Whole-Body Controller, $\boldsymbol{K}$, $\boldsymbol{W}_1$ and $\boldsymbol{W}_2$ were experimentally tuned so that an accurate tracking of the desired motion was achieved while avoiding too high joint velocities. The values selected for this purpose are $\boldsymbol{K}=diag\{1.0,1.0,1.0,0.1,0.1,0.1\}$, $\boldsymbol{W}_1=100\cdot diag\{10,10,10,5,5,5\}$ and $\boldsymbol{W}_2=3\cdot diag\{\boldsymbol{1}_m\}$. Moreover, in order to guarantee a locomotion behaviour (base following arm movements), $\boldsymbol{W}_3=diag\{\boldsymbol{0}_{n_b},\boldsymbol{1}_{n_a}\}$.

\begin{figure*}[t!]
    \centering
    {\includegraphics[width=0.85\textwidth, trim=4cm 0cm 3cm 0cm, clip=true]{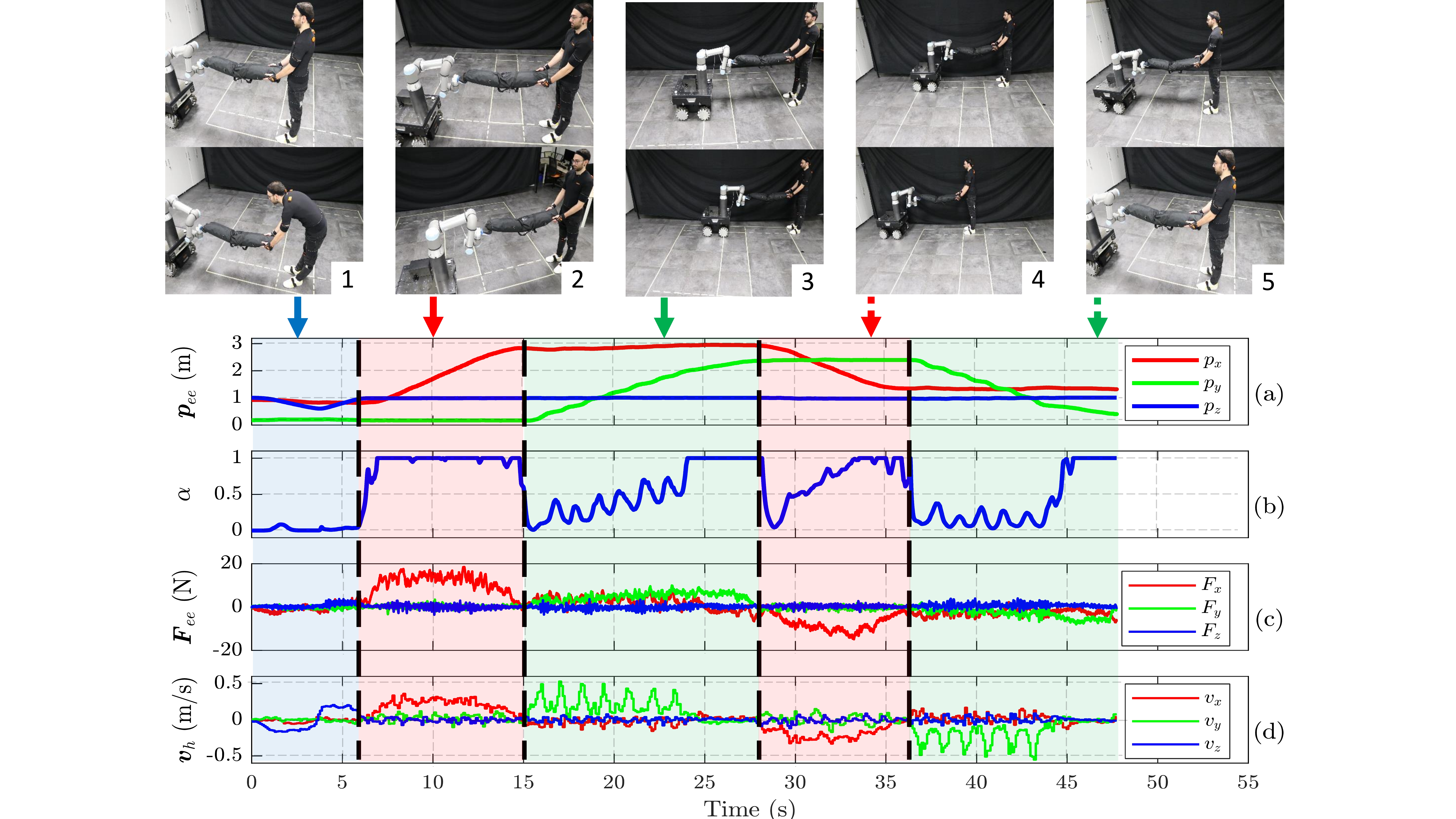}}
    \caption{Snapshots of the experiment: (1) Lowering and lifting; (2) Pulling; (3) Moving sideways to the right; (4) Pushing; and (5) Moving sideways to the left. Lower plots show the values obtained with the proposed controller during the collaborative transportation task of a typical trial.}
    \vspace{-0.68cm}
    \label{fig:intervals}
\end{figure*}

\subsection{Performance Metrics and Assessment Tools}
\label{performance_metrics_section}

After the experiments, the participants filled a questionnaire to rate different qualitative aspects of their experience for each controller. The questionnaire is composed of a standard part, namely the NASA-TLX \cite{HART1988139}, and a custom Likert scale-based part designed specifically for this study (see Table~\ref{custom}). Besides, for the quantitative evaluation of the task-related performance, the following metrics were used.

\begin{table}[b!]
\captionsetup{font=small}
    \centering
    \caption{Custom Questionnaire}
    \label{custom}
    \def\arraystretch{1.5}
    \begin{tabular}{|l l|}
    \hline
    \textbf{ID}  &\textbf{Question} \\
    \hline
    1 & I felt safe during the cooperation with the robot. \\
    \hline
    2 & The robot understood my intention easily. \\
    \hline
    3 & It was easy to keep the object inside the track.\\
    \hline
    4 & The robot made the co-carry comfortable for me. \\
    \hline
    5 & The robot's actions were harmless.  \\
     \hline
    6 & I was confident with the robot during the task. \\
     \hline
    7 & I would choose the robot as an assistant to accomplish the task.  \\
     \hline
    8 & The robot was helping me during the task. \\
     \hline
    9 & The robot was reliable during the task. \\
     \hline
    10 & The performance of the controller in different directions of the \\ 
    & co-carrying task was quite similar. \\
     \hline     
\end{tabular}
\end{table}

\begin{itemize}

    \item \textbf{Completion Time ($\boldsymbol{t_{c}}$)}: This is the elapsed time from the beginning to the end of the co-carrying task.
    
    \item \textbf{Alignment ($\boldsymbol{D_{AM}}$)}: This metric is formulated to calculate the robot performance in following human movements during the co-carrying task by taking into account the alignment of the object between human and robot. We assume that the experiment starts with an ideal object alignment (see Fig.~\ref{fig:system_overview_fig}). If the robot cannot keep the pace of the human during the task, the alignment of the object deviates from its ideal condition. In this situation, the distance between the optimal position of the robot's end-effector and its current position can be used as a metric to assess the performance of the robot. Thus, in a scenario where the robot perfectly follows the human, this metric will be 0. In order to track the object movements necessary for the calculation of this metric, we utilized the OptiTrack MoCap system. To this end, two different marker sets were placed near the human and robot grasping positions of the object. Based on these markers, the alignment metric is computed in the following way:

    \small
    \begin{equation}
    \begin{aligned}
        D_{AM}= \frac{\int_{t_{s}}^{t_{e}}{||\boldsymbol{r}_{crm}(t) - \boldsymbol{r}_{chm}(t) - (\boldsymbol{r}_{srm}- \boldsymbol{r}_{shm}) ||dt}}{t_{e}-t_{s}},
    \end{aligned}
    \end{equation}    
    \normalsize
    
    where $\boldsymbol{r}_{chm}$ and $\boldsymbol{r}_{crm}$ are the current human and robot marker positions, $\boldsymbol{r}_{shm}$ and $\boldsymbol{r}_{srm}$ are human and robot marker positions at the beginning of the experiment, and $t_s$ and $t_e$ indicate the starting and ending time of the experiment.

    \item \textbf{Effort ($\boldsymbol{E_{EMG}}$)}: In this study, the muscular activity of Anterior Deltoid (AD), Posterior Deltoid (PD), Biceps Brachii (BB), Triceps Brachii (TB), Flexor Carpi (FC), Extensor Digitorum (ED), Erector Spinae (ES) longissimus, and Multifidus (MF) was recorded using the Delsys Trigno platform, a wireless sEMG system commercialized by Delsys Inc. (Natick, MA, United States). The sEMG sensors were placed on the selected muscle groups of each participant's right arm and right part of the back according to SENIAM recommendations \cite{HERMENS2000}. Afterward, the signals obtained were filtered and normalized with their Maximum Voluntary Contractions (MVC). The effort metric is obtained for each individual muscle by calculating the average of this processed sEMG data during the experiment.
\end{itemize}

\subsection{Results}

For a deep analysis of the experimental results during the different movement directions, the experiment has been divided into 5 intervals as depicted in Fig.~\ref{fig:intervals}: (1) Lowering and lifting; (2) Pulling; (3) Moving sideways to the right; (4) Pushing; and (5) Moving sideways to the left. In the same figure, the results obtained from an example experiment with the proposed framework are shown.  

\begin{figure}[t!]
    \centering
    {\includegraphics[width=1\columnwidth, trim=7.1cm 9.8cm 16.5cm 1.6cm, clip=true]{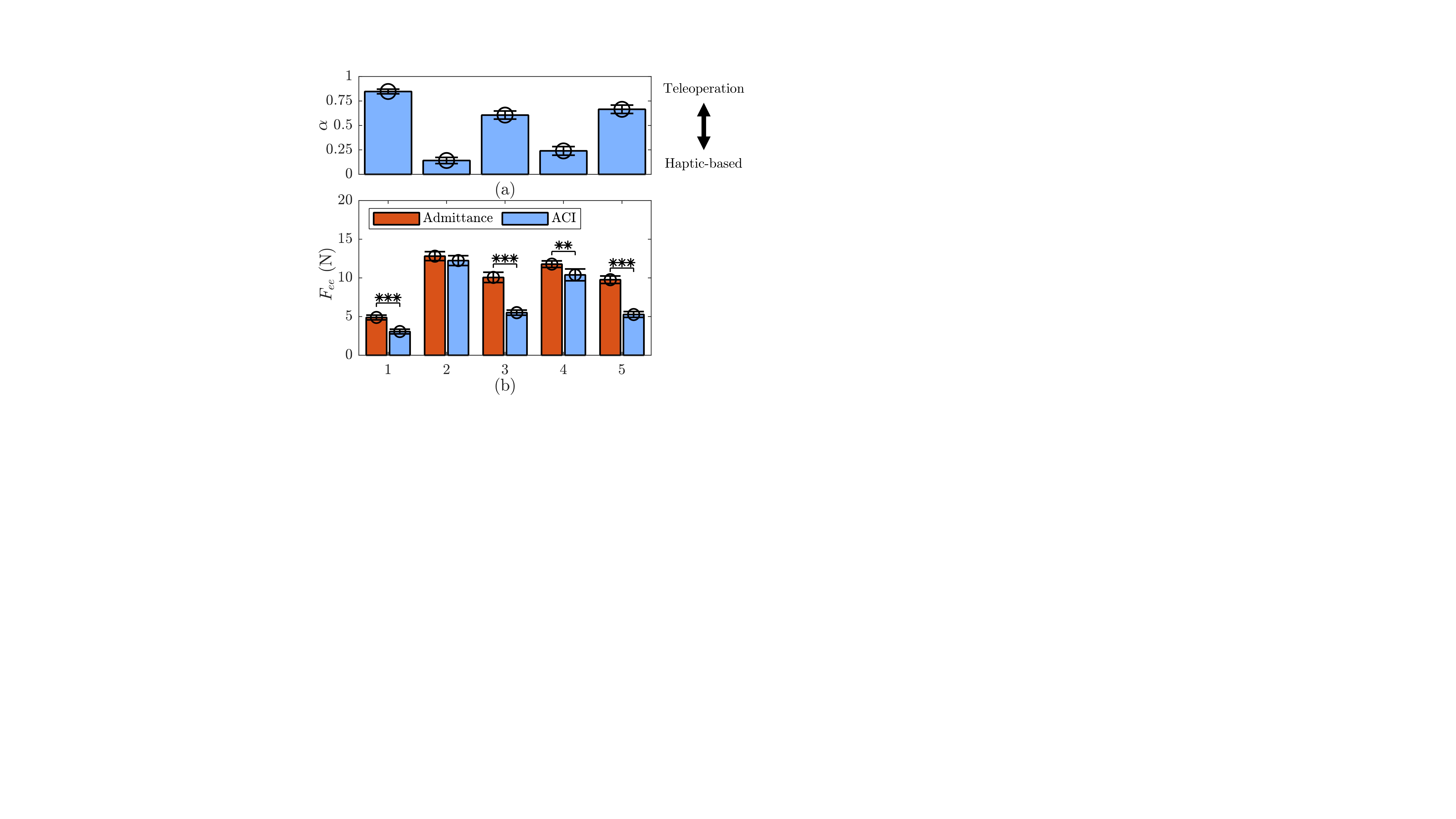}}
    \caption{(a) The means and standard errors of the adaptive index, $\alpha$, during the intervals, (b) the measured force amplitude from the end-effector of the robot for both controllers during the intervals along with outcomes of sign-test carried out: *: p $< 0.05$, **: p $< 0.01$, ***: p $< 0.001$.}
    \vspace{-0.4cm}

    \label{fig:interval_performance}
\end{figure}  

\begin{figure}[!b]
    \centering
    {\includegraphics[width=1\columnwidth, trim=0cm 0cm 0cm 0cm, clip=true]{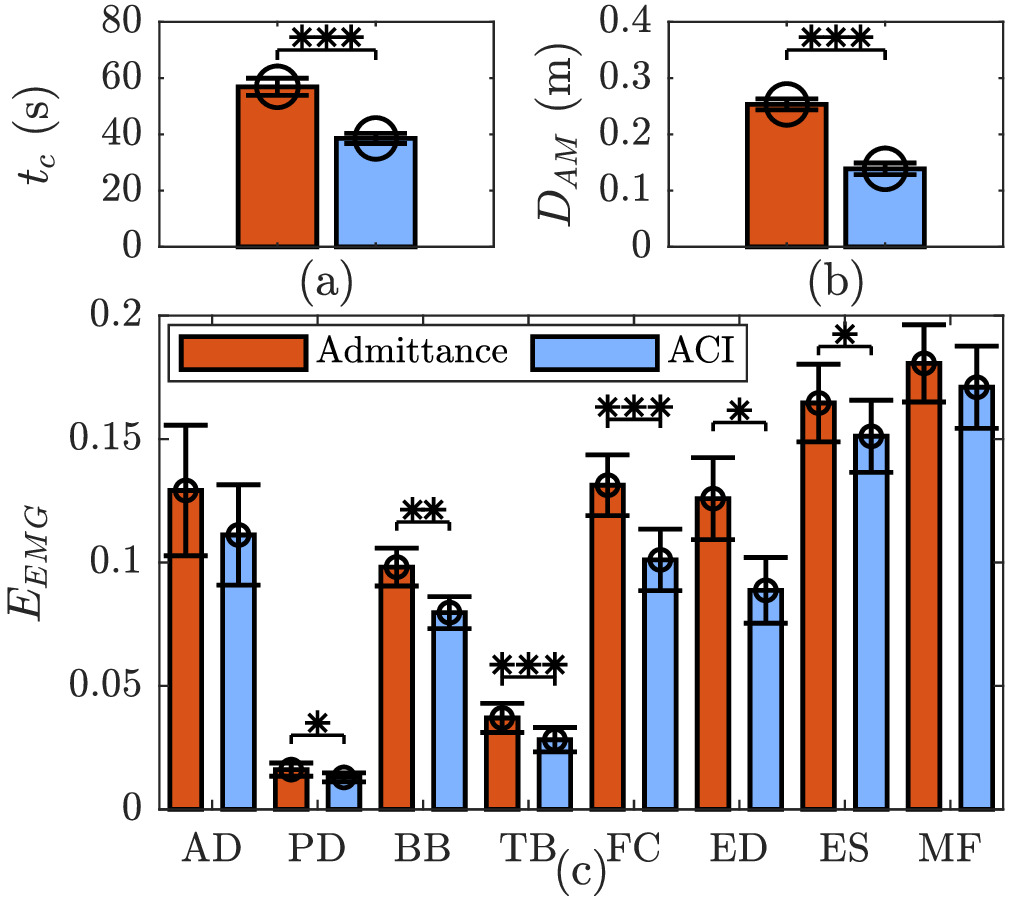}}
    \caption{The means and the standard errors for the performance metrics along with the outcomes of the sign-test carried out for the compared controllers: *: p $< 0.05$, **: p $< 0.01$, ***: p $< 0.001$.}
    \vspace{-0.0cm}
    \label{fig:performance_metrics}
\end{figure}

As mentioned in Sec.~\ref{sec:methodology}, when the adaptive index ($\alpha$) approaches 1, it shows that the object being carried is highly deformable. Vice-versa, when $\alpha$ approaches 0, this indicates the rigidity of the object. Fig.~\ref{fig:interval_performance}a shows the means and standard errors of $\alpha$ during the intervals, which are calculated by averaging for all participants. As it can be seen, the highest mean is observed during the first interval, where it is close to 1. Conversely, the smallest one is observed during the second one among all intervals, where it approaches 0. Moreover, the third, fifth and fourth intervals exhibit intermediate values
, with means of 0.6 for the former two and 0.25 for the latter. These results indicate that the object in combination with the grasping type used, is not rigid while lowering and lifting, and it behaves as a non-deformable object when it is fully stretched while pulling. Additionally, the object is partially deformable when it is pushed and moved sideways. Note that, similar deformability is estimated in the two directions of sideways movements. These outcomes are expected considering the characteristics of the object being carried combined with the non-purely rigid Pisa/IIT Softhand placed at the end-effector.

\begin{figure}[t!]
    \centering
    {\includegraphics[width=1\columnwidth, trim=0cm 0cm 0cm 0cm, clip=true]{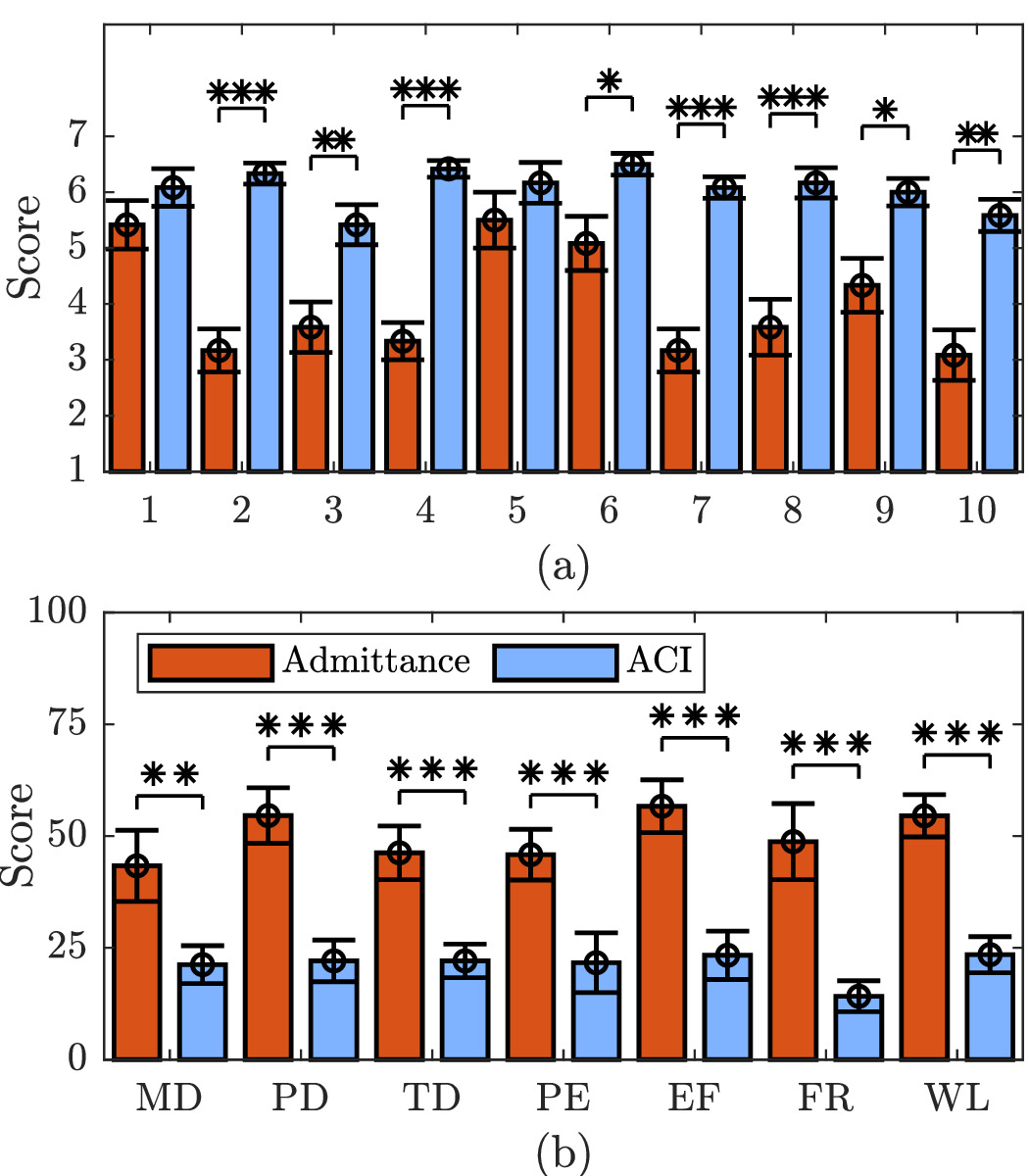}}
    \caption{The means and the standard errors obtained from (a) the custom and (b) the NASA-TLX questionnaires for the two controllers along with the outcomes of the sign-test carried out for the compared controllers: *: p $< 0.05$, **: p $< 0.01$, ***: p $< 0.001$.}
    \vspace{-0.4cm}
    \label{fig:questionnaires}
\end{figure} 

The means and standard errors of the force amplitude measured at the robot's end-effector for the two controllers during the intervals are depicted in Fig.~\ref{fig:interval_performance}b. These results are calculated by finding their average over all participants. When these results are analyzed, it can be seen that the admittance controller presents statistical significant higher force amplitudes according to sign-test for all intervals except the second one. This difference is due to the deformability of the object, which is indicated by high values of $\alpha$. These high values can be interpreted as the difficulty of transferring forces from the human to the robot through the object. The ACI copes with this issue by estimating humans intention by means of the velocity of their hand. Instead, the admittance controller fails, and the cooperating human needs to deform the object to allow the force transmission and the intention communication. In contrast, in the second interval no significant differences are revealed between the two controllers, as expected. Indeed, in that segment of the task the object is almost purely rigid (i.e. $\alpha$ approaches 1).

Fig.~\ref{fig:performance_metrics} illustrates the means and standard errors of the performance metrics described in subsection~\ref{performance_metrics_section} for both the proposed controller and the admittance controller. The performance metrics are obtained by averaging these measured quantities for all participants. Sign-tests are performed to evaluate the effect of the controller on the performance metrics and the outcomes of them can be found (see Fig.~\ref{fig:performance_metrics}). The results show that the completion time and the alignment for the proposed controller are significantly lower (p $< 0.001$) than those of the admittance controller (see Fig.~\ref{fig:performance_metrics}a, \ref{fig:performance_metrics}b). The mean efforts for all the selected muscle groups are lower under the proposed controller (see Fig.~\ref{fig:performance_metrics}c). However, there is no significant effect of controller type on the effort of AD and MF muscles. These results indicate that participants completed the task faster, with less effort and higher effectiveness under the proposed controller.

The qualitative results are reported in Fig.~\ref{fig:questionnaires}. For each questionnaire statement, the means and standard errors for both controllers are represented as bar plots, and the outcomes of the sign-tests are reported. Particularly, Fig.~\ref{fig:questionnaires}a depicts the outcomes of the custom questionnaire, which aims to obtain a subjective evaluation of the two controllers, while Fig.~\ref{fig:questionnaires}b shows the results obtained for the NASA-TLX questionnaire through which the participants score their perceived workload for six subscales. The statements of the custom questionnaire can be classified into 3 categories: safety (Q.1 and 5); performance (Q.2, 3, 8, and 10); and usability (Q.4, 6, 7, and 9). As shown in Fig.~\ref{fig:questionnaires}a, the proposed controller has better score means compared with the admittance controller in all statements of the custom questionnaire. In addition, statistically significant differences (p $< 0.05$) are observed for all statements apart from the ones related to safety. These results demonstrate that the proposed controller surpasses the admittance controller in terms of performance and usability according to the participants while ensuring safety which is of utmost importance for a controller in pHRI. Regarding the NASA-TLX questionnaire (see Fig.~\ref{fig:questionnaires}b), the ACI has statistically significant superior scores in all the categories, as well as in the computed overall workload.

\begin{figure*}[t!]
    \centering
    \resizebox{0.98\textwidth}{!}{\rotatebox{0}{\includegraphics[trim=0cm 0cm 0cm 0cm, clip=true]{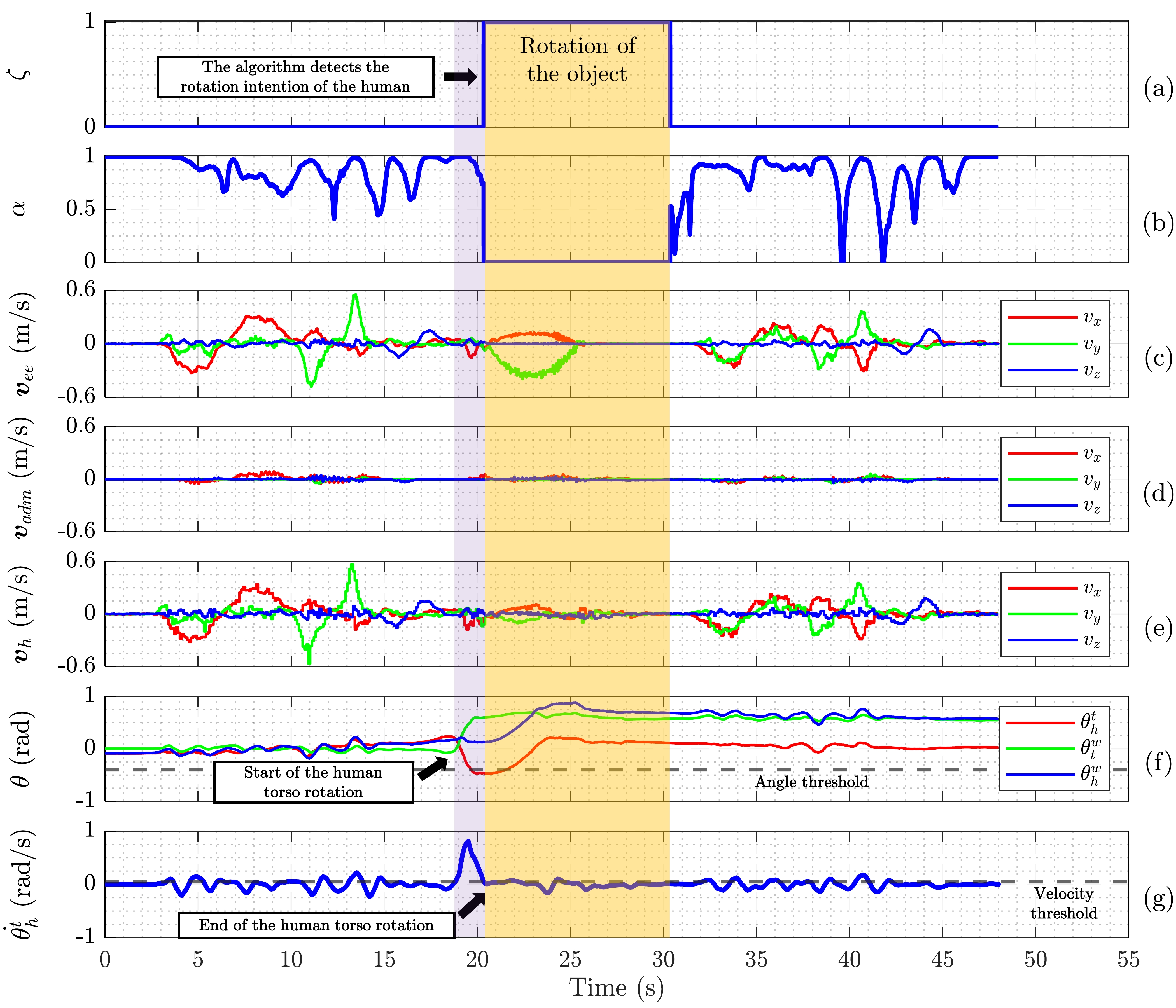}}}
    \caption{The results obtained during the showcase where a manikin is collaborative carried through a sheet. The graphs show (a) the human rotation intention $\zeta$ detected by the algorithm, (b) the adaptive index $\alpha$, (c) the end-effector velocity of the robot $\boldsymbol{{v}}_{ee}$, (d) the admittance reference velocity $\boldsymbol{{v}}_{adm}$, (e) the human hand velocity $\boldsymbol{{v}}_{h}$, (f) the relative yaw angle between the human torso and the hand $\theta_{h}^{t}$, the torso angle with respect to the world frame $\theta_{t}^{w}$, the torso angle with respect to the world frame $\theta_{h}^{w}$, and the selected angle threshold of 0.4 rad (the dashed line) for the algorithm, and (g) the torso yaw velocity $\dot{\theta_{t}^{w}}$ and the selected velocity threshold of 0.05 rad/s (the dashed line) for the algorithm.}
    \label{fig:practical_showcase}
    \vspace{-0.5cm}
\end{figure*}

\section{PRACTICAL SHOWCASE}
\label{sec:practical_showcase}

Previously, the applicability of the ACI has been demonstrated in experimental settings, i.e., the user-study and the extremities experiments. In the latter, objects were straightforward to model, while in the former the object was hard to characterize even though anticipating its behaviour and foreseeing the results in terms of deformability was still feasible. In this section, we have validated the proposed controller in a more realistic and challenging scenario including rotational movements, where outcomes in terms of object deformability are tough to predict. In particular, a 4kg manikin laying on a sheet is carried by grasping the sheet's ends. Note that, the complexity of its deformability properties is mostly due to the manikin's freedom to move inside the sheet, which changes the load distribution during co-transportation. A video of this showcase can be found in the multimedia attachment\footnote{Find the video also here \url{https://youtu.be/oyfUkYj5WYw}}.

Fig.~\ref{fig:practical_showcase} depicts the data associated to this experiment. The human operator began the co-carry task with the back-and-forth movements in all 3 Cartesian axes (X-Y-Z order). Apart from the sudden drops during the change of directions, $\alpha$ was closer to 1 along all the axes (see Fig.~\ref{fig:practical_showcase}b). Then, the human operator started to turn his torso in order to rotate the robot to an intended angle (see Fig.~\ref{fig:practical_showcase}f). This motion caused a relative yaw angle between the human torso and hand ($\theta_{h}^{t}$) which was generated by human torso rotation ($\theta_{t}^{w}$). When the torso yaw velocity decreased below the threshold of 0.05 rad/s (see Fig.~\ref{fig:practical_showcase}g), which shows the intended torso rotation angle was reached and human decided to stop, Alg.~\ref{alg:cap} detected the human rotation intention (see Fig.~\ref{fig:practical_showcase}a). Later, the robot started to move from its current pose to the calculated desired pose ($\boldsymbol{T}_{r,des}^{w}$) according to this detected torso angle. After the robot reached its planned pose, the human operator finished the co-carry task by transporting the object in all 3 directions as in the first part.

\section{DISCUSSION and CONCLUSION}
\label{sec:discussion}

Collaboration between a human and a robot can be an effective solution for the task of carrying an object that can not be performed by a single partner. In such situations, we believe that the pHRI frameworks that enable transporting objects with different deformability behaviours are missing in the state-of-the-art. This paper presented an adaptive control framework that can handle objects with unknown deformability for human-robot collaborative transportation. Instead of relying just on the deficient haptic information that is transferred through the deformable object, the proposed controller allows us to benefit from the information of human movements. Based on this perceived sensory information, it is also possible to react to changes in deformability of the object through an adaptive index computed online. Moreover, the proposed control framework is able to plan the motion of the mobile manipulator when it detects the rotation intention of the human by using the torso and hand movements. 

In this study, the effectiveness of the presented framework was evaluated in the extremities of the object deformability range with a purely rigid aluminum profile and a highly deformable rope. After validating that our framework was successful in handling these extremities, we compared its performance to an admittance controller during a collaborative carrying task of a partially deformable object. A thorough analysis based on quantitative and qualitative results of the 12-subjects user study was carried out to assess the performance of the two controllers. The results showed that the proposed controller outperforms than the admittance controller on providing assistance during co-carrying of deformable objects. In addition to this user study, we demonstrated the usability of our framework during a more realistic scenario in which the deformation of the object is complicated to estimate beforehand.

In the future, we plan to make the proposed framework more applicable for the industrial scenarios by replacing our human motion capture system (Xsens suit) with a less expensive vision-based system. Especially, placing the cameras of this vision-based system on the robot will make the framework flexible to use and overcome the disadvantages of Xsens such as the need of wearing a suit.

\bibliographystyle{ieeetr}
\bibliography{biblio.bib}

\begin{thebibliography}{10}

\bibitem{krueger2017have}
A.~B. Krueger, ``Where have all the workers gone? an inquiry into the decline
  of the us labor force participation rate,'' {\em Brookings papers on economic
  activity}, vol.~2017, no.~2, p.~1, 2017.

\bibitem{dework}
J.~De~Kok, P.~Vroonhof, J.~Snijders, G.~Roullis, M.~Clarke, K.~Peereboom,
  P.~van Dorst, and I.~Isusi, {\em Work-related musculoskeletal disorders:
  prevalence, costs and demographics in the EU}.
\newblock Publications Office, 2020.

\bibitem{kim2019adaptable}
W.~Kim, M.~Lorenzini, P.~Balatti, P.~D. Nguyen, U.~Pattacini, V.~Tikhanoff,
  L.~Peternel, C.~Fantacci, L.~Natale, G.~Metta, {\em et~al.}, ``Adaptable
  workstations for human-robot collaboration: A reconfigurable framework for
  improving worker ergonomics and productivity,'' {\em IEEE Robotics \&
  Automation Magazine}, vol.~26, no.~3, pp.~14--26, 2019.

\bibitem{Arash}
A.~Ajoudani, A.~M. Zanchettin, S.~Ivaldi, A.~Albu-Schäffer, K.~Kosuge, and
  O.~Khatib, ``Progress and prospects of the human-robot collaboration,'' {\em
  Autonomous Robots}, vol.~42, pp.~957--975, Oct. 2018.

\bibitem{gandarias2022enhancing}
J.~M. Gandarias, P.~Balatti, E.~Lamon, M.~Lorenzini, and A.~Ajoudani,
  ``Enhancing flexibility and adaptability in conjoined human-robot industrial
  tasks with a minimalist physical interface,'' in {\em 2022 Proceedings of
  IEEE International Conference on Robotics and Automation (ICRA)}, IEEE, 2022.

\bibitem{mortl2012role}
A.~M{\"o}rtl, M.~Lawitzky, A.~Kucukyilmaz, M.~Sezgin, C.~Basdogan, and
  S.~Hirche, ``The role of roles: Physical cooperation between humans and
  robots,'' {\em The International Journal of Robotics Research}, vol.~31,
  no.~13, pp.~1656--1674, 2012.

\bibitem{dumora2012experimental}
J.~Dumora, F.~Geffard, C.~Bidard, T.~Brouillet, and P.~Fraisse, ``Experimental
  study on haptic communication of a human in a shared human-robot
  collaborative task,'' in {\em 2012 IEEE/RSJ International Conference on
  Intelligent Robots and Systems}, pp.~5137--5144, IEEE, 2012.

\bibitem{Takubo2002}
T.~Takubo, H.~Arai, Y.~Hayashibara, and K.~Tanie, ``Human-robot cooperative
  manipulation using a virtual nonholonomic constraint,'' {\em The
  International Journal of Robotics Research}, vol.~21, no.~5-6, pp.~541--553,
  2002.

\bibitem{doganay2022}
D.~Sirintuna, A.~Giammarino, and A.~Ajoudani, ``Human-robot collaborative
  carrying of objects with unknown deformation characteristics,'' in {\em 2022
  IEEE/RSJ International Conference on Intelligent Robots and Systems (IROS)},
  IEEE, 2022.

\bibitem{ikeura1995variable}
R.~Ikeura and H.~Inooka, ``Variable impedance control of a robot for
  cooperation with a human,'' in {\em Proceedings of 1995 IEEE International
  Conference on Robotics and Automation}, vol.~3, pp.~3097--3102, IEEE, 1995.

\bibitem{Duchaine}
V.~Duchaine and C.~M. Gosselin, ``General model of human-robot cooperation
  using a novel velocity based variable impedance control,'' in {\em Second
  Joint EuroHaptics Conference and Symposium on Haptic Interfaces for Virtual
  Environment and Teleoperator Systems (WHC'07)}, pp.~446--451, 2007.

\bibitem{duchaine2009safe}
V.~Duchaine and C.~Gosselin, ``Safe, stable and intuitive control for physical
  human-robot interaction,'' in {\em 2009 IEEE International Conference on
  Robotics and Automation}, pp.~3383--3388, IEEE, 2009.

\bibitem{lecours2012variable}
A.~Lecours, B.~Mayer-St-Onge, and C.~Gosselin, ``Variable admittance control of
  a four-degree-of-freedom intelligent assist device,'' in {\em 2012 IEEE
  international conference on robotics and automation}, pp.~3903--3908, IEEE,
  2012.

\bibitem{evrard2009teaching}
P.~Evrard, E.~Gribovskaya, S.~Calinon, A.~Billard, and A.~Kheddar, ``Teaching
  physical collaborative tasks: Object-lifting case study with a humanoid,'' in
  {\em 2009 9th IEEE-RAS International Conference on Humanoid Robots},
  pp.~399--404, IEEE, 2009.

\bibitem{bussy2012proactive}
A.~Bussy, P.~Gergondet, A.~Kheddar, F.~Keith, and A.~Crosnier, ``Proactive
  behavior of a humanoid robot in a haptic transportation task with a human
  partner,'' in {\em 2012 IEEE RO-MAN: The 21st IEEE International Symposium on
  Robot and Human Interactive Communication}, pp.~962--967, IEEE, 2012.

\bibitem{agravante}
D.~J. Agravante, A.~Cherubini, A.~Bussy, P.~Gergondet, and A.~Kheddar,
  ``Collaborative human-humanoid carrying using vision and haptic sensing,'' in
  {\em 2014 IEEE International Conference on Robotics and Automation (ICRA)},
  pp.~607--612, 2014.

\bibitem{xinbo}
X.~Yu, W.~He, Q.~Li, Y.~Li, and B.~Li, ``Human-robot co-carrying using visual
  and force sensing,'' {\em IEEE Transactions on Industrial Electronics},
  vol.~68, no.~9, pp.~8657--8666, 2021.

\bibitem{Karayiannidis}
Y.~Karayiannidis, C.~Smith, and D.~Kragic, ``Mapping human intentions to robot
  motions via physical interaction through a jointly-held object,'' in {\em The
  23rd IEEE International Symposium on Robot and Human Interactive
  Communication}, pp.~391--397, 2014.

\bibitem{sanchez2018robotic}
J.~Sanchez, J.-A. Corrales, B.-C. Bouzgarrou, and Y.~Mezouar, ``Robotic
  manipulation and sensing of deformable objects in domestic and industrial
  applications: a survey,'' {\em The International Journal of Robotics
  Research}, vol.~37, no.~7, pp.~688--716, 2018.

\bibitem{Maeda}
Y.~Maeda, T.~Hara, and T.~Arai, ``Human-robot cooperative manipulation with
  motion estimation,'' in {\em Proceedings 2001 IEEE/RSJ International
  Conference on Intelligent Robots and Systems. Expanding the Societal Role of
  Robotics in the the Next Millennium (Cat. No.01CH37180)}, vol.~4,
  pp.~2240--2245 vol.4, 2001.

\bibitem{jerk_model}
S.~Miossec and A.~Kheddar, ``Human motion in cooperative tasks: Moving object
  case study,'' in {\em 2008 IEEE International Conference on Robotics and
  Biomimetics}, pp.~1509--1514, 2009.

\bibitem{kruse2015collaborative}
D.~Kruse, R.~J. Radke, and J.~T. Wen, ``Collaborative human-robot manipulation
  of highly deformable materials,'' in {\em 2015 IEEE international conference
  on robotics and automation (ICRA)}, pp.~3782--3787, IEEE, 2015.

\bibitem{delpreto2019sharing}
J.~DelPreto and D.~Rus, ``Sharing the load: Human-robot team lifting using
  muscle activity,'' in {\em 2019 International Conference on Robotics and
  Automation (ICRA)}, pp.~7906--7912, IEEE, 2019.

\bibitem{seraji1990improved}
H.~Seraji and R.~Colbaugh, ``Improved configuration control for redundant
  robots,'' {\em Journal of Robotic Systems}, vol.~7, no.~6, pp.~897--928,
  1990.

\bibitem{chiaverini1992weighted}
S.~Chiaverini, O.~Egeland, and R.~K. Kanestrom, ``Weighted damped least-squares
  in kinematic control of robotic manipulators,'' {\em Advanced robotics},
  vol.~7, no.~3, pp.~201--218, 1992.

\bibitem{deo1995overview}
A.~S. Deo and I.~D. Walker, ``Overview of damped least-squares methods for
  inverse kinematics of robot manipulators,'' {\em Journal of Intelligent and
  Robotic Systems}, vol.~14, no.~1, pp.~43--68, 1995.

\bibitem{hollerbach1987redundancy}
J.~Hollerbach and K.~Suh, ``Redundancy resolution of manipulators through
  torque optimization,'' {\em IEEE Journal on Robotics and Automation}, vol.~3,
  no.~4, pp.~308--316, 1987.

\bibitem{wampler1986manipulator}
C.~W. Wampler, ``Manipulator inverse kinematic solutions based on vector
  formulations and damped least-squares methods,'' {\em IEEE Transactions on
  Systems, Man, and Cybernetics}, vol.~16, no.~1, pp.~93--101, 1986.

\bibitem{nakanishi2005comparative}
J.~Nakanishi, R.~Cory, M.~Mistry, J.~Peters, and S.~Schaal, ``Comparative
  experiments on task space control with redundancy resolution,'' in {\em 2005
  IEEE/RSJ International Conference on Intelligent Robots and Systems},
  pp.~3901--3908, 2005.

\bibitem{wu2021unified}
Y.~Wu, E.~Lamon, F.~Zhao, W.~Kim, and A.~Ajoudani, ``Unified approach for
  hybrid motion control of moca based on weighted whole-body cartesian
  impedance formulation,'' {\em IEEE Robotics and Automation Letters}, vol.~6,
  no.~2, pp.~3505--3512, 2021.

\bibitem{HART1988139}
S.~G. Hart and L.~E. Staveland, ``Development of nasa-tlx (task load index):
  Results of empirical and theoretical research,'' in {\em Human Mental
  Workload} (P.~A. Hancock and N.~Meshkati, eds.), vol.~52 of {\em Advances in
  Psychology}, pp.~139--183, North-Holland, 1988.

\bibitem{HERMENS2000}
H.~J. Hermens, B.~Freriks, C.~Disselhorst-Klug, and G.~Rau, ``Development of
  recommendations for semg sensors and sensor placement procedures,'' {\em
  Journal of Electromyography and Kinesiology}, vol.~10, pp.~361 -- 374, Oct.
  2000.

\end{thebibliography}

\begin{IEEEbiography}[{\includegraphics[width=1in,height=1.25in,clip,keepaspectratio,trim=6cm 13cm 8cm 3cm]{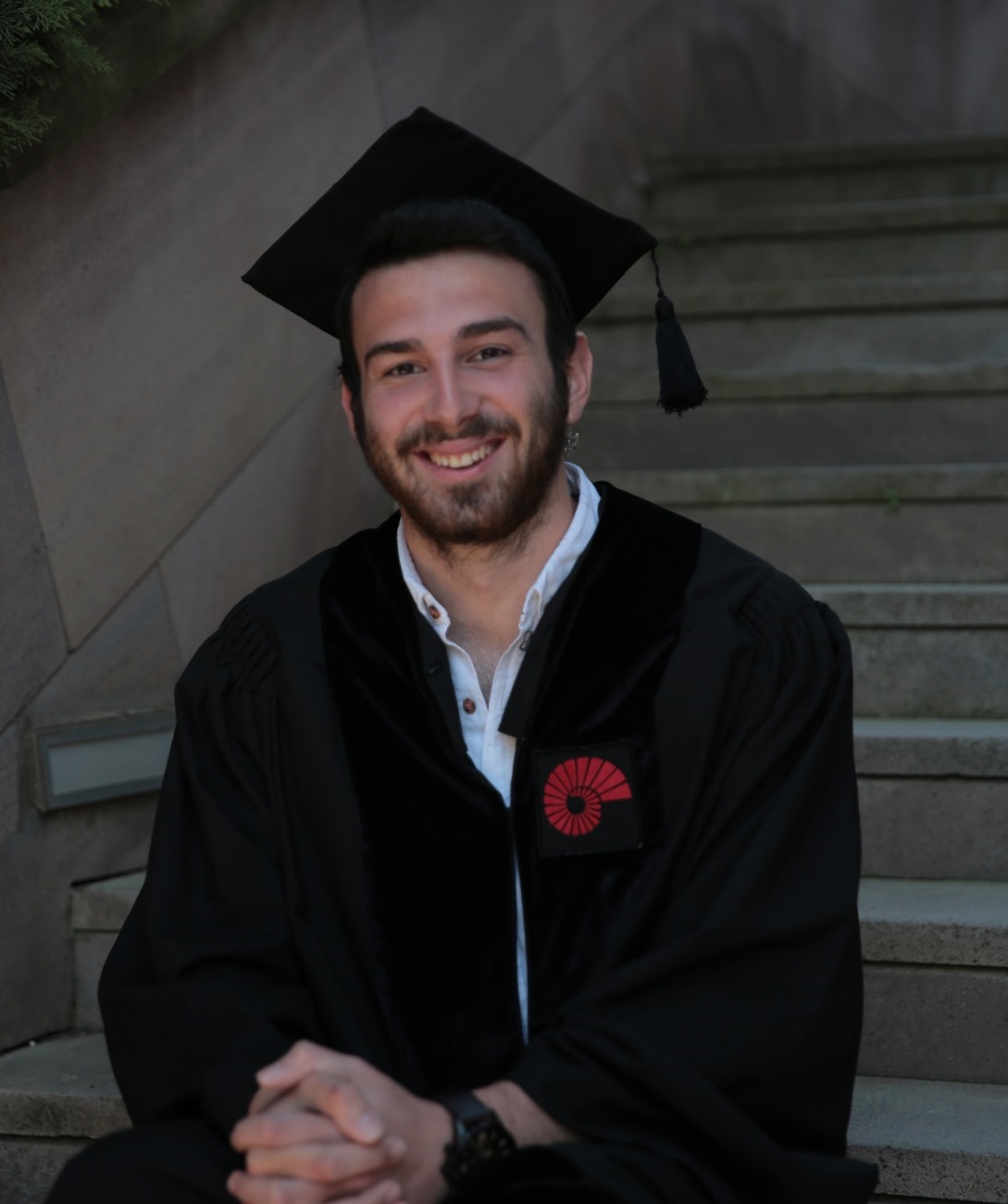}}] 
{Doganay Sirintuna} is a fellow researcher at the Human-Robot Interfaces and Physical Interaction Lab (HRI$^2$) at the Italian Institute of Technology (IIT). He received his B.Sc. and M.Sc. degrees in Mechanical Engineering from Koç University, Istanbul, in 2018 and 2020, respectively. He was granted the Academic Excellence Award at his graduation from Koç University in 2020. His research interests include physical human-robot interaction, mobile manipulation, and robot learning.
\end{IEEEbiography}

\begin{IEEEbiography}[{\includegraphics[width=1in,height=1.25in,clip,keepaspectratio]{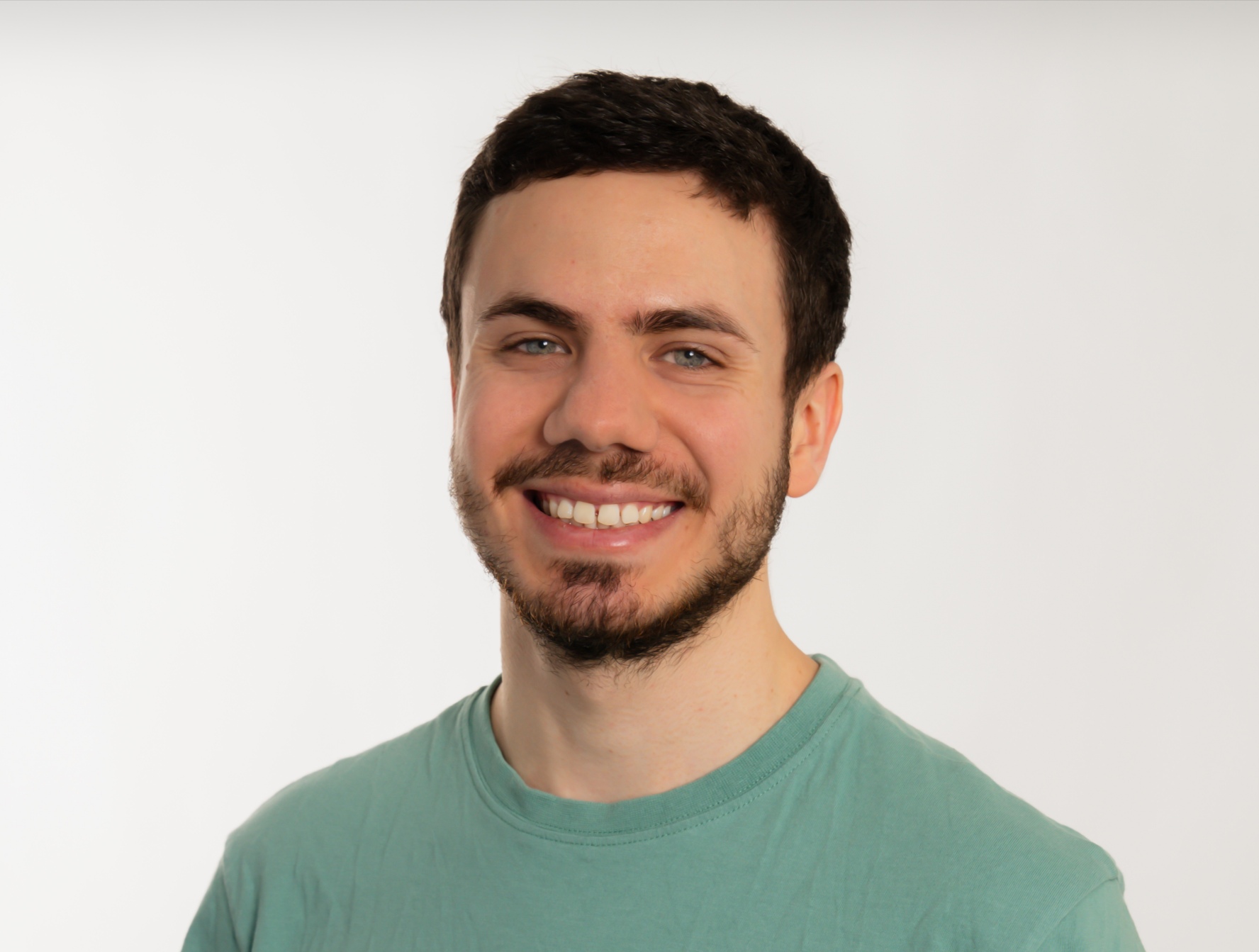}}]{Alberto Giammarino}
is a fellow researcher at the Human-Robot Interfaces and Physical Interaction Lab (HRI$^2$) at the Italian Institute of Technology (IIT). He received the B.S. in Mechanical Engineering from the University of Bologna in 2018, the M.S. in Mechanical Engineering with a major in Mechatronics and Robotics from Politecnico di Milano in 2021, and spent half of his M.S. at ETH Z\"urich. His current research interests include physical Human-Robot Interaction, Optimal Control and Reinforcement Learning.
\end{IEEEbiography}

\begin{IEEEbiography}[{\includegraphics[width=1in,height=1.25in,clip,keepaspectratio]{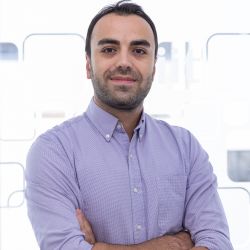}}]{Arash Ajoudani} is a tenured senior scientist at the Italian Institute of Technology (IIT), where he leads the Human-Robot Interfaces and physical Interaction (HRI$^2$) laboratory. He received his PhD degree in Robotics and Automation from University of Pisa and IIT in 2014. He is a recipient of the European Research Council (ERC) starting grant 2019, the coordinator of the Horizon-2020 project SOPHIA, and the co-coordinator of the Horizon-2020 project CONCERT. His main research interests are in physical human-robot interaction, mobile manipulation, robust and adaptive control, assistive robotics, and tele-robotics.
\end{IEEEbiography}

\end{document}